\documentclass[review]{elsarticle}
\usepackage{hyperref}

\journal{Journal of \LaTeX\ Templates}

\newcommand{\argmin}{\mathop{\rm arg~min}\limits}

\begin{document}

\begin{frontmatter}
\title{Landmark Map: An Extension of the Self-Organizing Map 
for a User-Intended Nonlinear Projection}

\author{Akinari Onishi\fnref{myfootnote}}
\address{1-33 Yayoi-cho, Inage-ku, Chiba, Japan}
\fntext[myfootnote]{Since 2017.}

\author[]{Center for Frontier Medical Engineering, Chiba University}
\ead[url]{http://onishi.starfree.jp/eng.html}

\cortext[mycorrespondingauthor]{Corresponding author}
\ead{a-onishi@chiba-u.jp}


\begin{abstract}
The self-organizing map (SOM) is an unsupervised 
artificial neural network that is widely 
used in, e.g., data mining and visualization. 
Supervised and semi-supervised learning methods have been proposed for the SOM. 
However, their teacher labels do not describe the relationship 
between the data and the location of nodes. 
This study proposes a landmark map (LAMA), 
which is an extension of the SOM that utilizes several landmarks, e.g., pairs of nodes and data points. 
LAMA is designed to obtain a user-intended nonlinear projection 
to achieve, e.g., the landmark-oriented data visualization.  
To reveal the learning properties of LAMA, 
the Zoo dataset from the UCI Machine Learning Repository and an artificial formant dataset were analyzed. 
The analysis results of the Zoo dataset indicated that 
LAMA could provide a new data view such as 
the landmark-centered data visualization. 
Furthermore, the artificial formant data analysis 
revealed that LAMA successfully provided 
the intended nonlinear projection 
associating articular movement with  
vertical and horizontal movement of a computer cursor. 
Potential applications of LAMA include data mining, recommendation systems, 
and human-computer interaction. 
\end{abstract}

\end{frontmatter}


\section{Introduction}

\noindent
The self-organizing map (SOM), proposed by Teuvo~Kohonen, is a type of artificial neural network that 
provides a nonlinear projection from a high-dimensional space 
to a low-dimensional discrete space \cite{kohonen2001, Kohonen2013}. 
The SOM is trained by unique unsupervised learning using its own architecture of nodes, 
which is also referred to as the topology-preserving mapping algorithm \cite{kraaijveld1995nonlinear, Uriarte2005}. 
It is often used to visualize relationships of data 
via a two-dimensional contour plot called the unified distance matrix (U-matrix) \cite{ultsch1990kohonen, Ultsch2003}. 
Owing to its convenience of visualization, 
the SOM has been used for data mining, especially for clustering \cite{Vesanto2000}. 
For example, it has been used for the clustering of gait kinematics \cite{Caldas2018}, 
farm profitability \cite{Sulkava2015}, and groundwater quality \cite{Belkhiri2018}. 
In addition, SOMs have been applied to classification \cite{Ultsch2005}. 
For instance, they have been used to classify electroencephalography (EEG) signals 
in applications of human-computer interaction (HCI) such a brain-computer interface (BCI) \cite{Al-Ketbi2013}. 
Furthermore, SOMs have also been utilized for facial recognition \cite{Lawrence1997}, 
hand and body tracking \cite{Coleca2015}, gesture recognition \cite{Chen2016}, 
fault detection \cite{Yu2015},
mental task classification of EEG for BCI \cite{Liu2005}, 
EEG-based emotion recognition \cite{Khosrowabadi2010}, 
emotion recognition using geometric facial features \cite{Majumder2014}, 
and hazard detection of motorcycles \cite{Selmanaj2017}. 
Although SOM is useful for mining unlabeled data because of its unsupervised learning, 
they do not take advantage of labeled or partly labeled data in supervised or semi-supervised learning.

Some supervised and semi-supervised learning methods have been proposed for SOMs. 
In early studies, the class label was concatenated 
to the input data \cite{kohonen1988neural,fessant2001comparison, goren2000supervised}.
Kohonen et al.~adopted learning vector quantization (LVQ), 
which updates only the winner node, and then fine-tuned the SOM \cite{Kohonen1998}. 
Hagenbuchner et al.~proposed a supervised SOM that assigns labels to neurons 
and then rejects the projected data if it has a different label \cite{Hagenbuchner2005}. 
Shen et al.~proposed a three-layer semi-supervised learning method for SOMs \cite{Shen2010}.
The method first trains SOM on both labeled and unlabeled data. 
Next, it labels nodes and trains a classifier using labeled data. 
Herrmann et al.~introduced Zhu's label propagation method for the SOM \cite{Herrmann2007}.
Thus, regardless of the relationships between the nodes and the data,  
many supervised and semi-supervised learning algorithms using class labels have been proposed for the SOM. 
However, these methods do not focus on clarifying the relationship 
between the original data and the location of projected nodes.
If we have a clear hypothesis or objective for the projection, 
the data and nodes of the SOM can be associated as teachers during learning.

To obtain a user-intended nonlinear projection, 
this study proposes a {\it landmark map} (LAMA), 
which is an extension of the SOM learned by {\it landmarks}, i.e., several pairs of 
data and associated nodes.
Such landmarks are implemented as landmark nodes having landmark data. 
By an alternating update method using the given data and landmark data, 
the LAMA is trained so that the data projected onto the landmark nodes approach the landmark data, 
while the properties of the SOM are retained as far as possible. 
To reveal the difference between the learning properties of the LAMA and the SOM, 
they are evaluated using two datasets, 
namely the Zoo dataset (UCI Machine Learning Repository, University of California, Irvine, CA, US) and an artificial formant dataset.

The LAMA differs from previously proposed supervised and semi-supervised SOM methods 
in that it uses several landmarks instead of teacher labels. 
This implies that the LAMA directly combines 
a small number of data and nodes as teachers without using traditional class labels.
In addition, the LAMA is designed to obtain 
a user-intended nonlinear projection 
by setting several landmark nodes. 
On the other hand, the previously proposed supervised and semi-supervised SOM 
aims for classification, clustering, and dimensionality reduction. 
The LAMA is not designed to address those traditional 
supervised or unsupervised learning problems 
but addresses the newly defined problem. 
Due to the differences in the objectives and problem settings,  
a comparison between LAMA and the proposed supervised or 
semi-supervised SOM is not straightforward. 
In this study, LAMA was compared with the unsupervised online SOM 
in terms of learning properties seen mainly in the input space.

\section{Landmark map}

\subsection{Artificial neural network architecture of LAMA}

\noindent
LAMA is a type of artificial neural network that 
projects input data from a continuous input space to a discrete output space. 
LAMA learns its projection preserving arrangement of nodes in the output space (topology) 
and retains relationships between the landmark data in the input space and the landmark nodes in the output space.  
In other words, LAMA is an extension of SOMs with several landmarks that indicate connections 
between the {\it landmark data} and the {\it landmark nodes}. 
LAMA can be applied to obtain a nonlinear projection 
from higher-dimensional space to a lower-dimensional output space. 
In addition, it can provide a modified nonlinear projection 
between the same-dimensional input and output space.

\begin{figure}[!h]
\includegraphics[width=\linewidth]{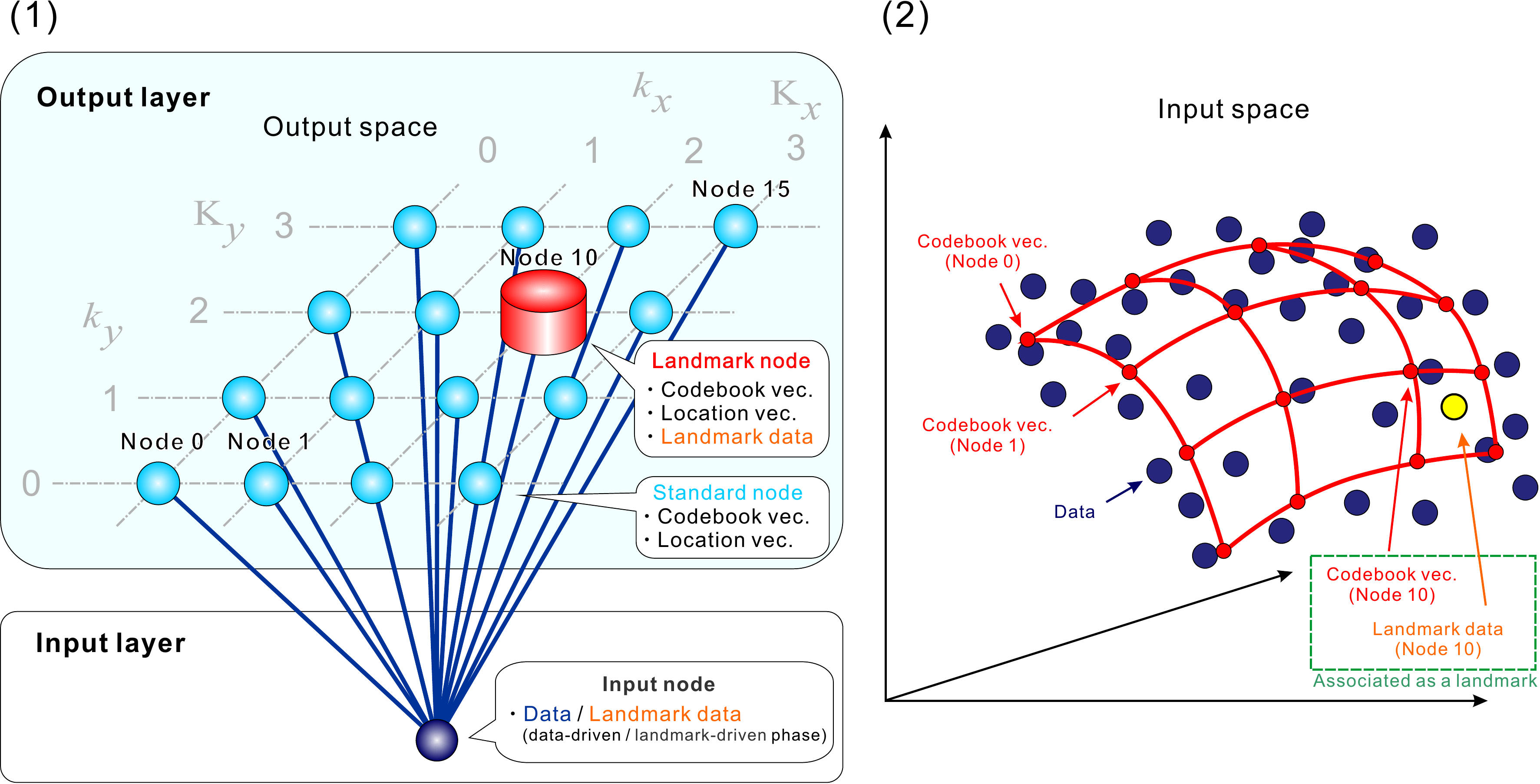}
\caption{Overview of the landmark map (LAMA). 
(1) {\it Artificial neural network architecture of the LAMA}. 
This consists of standard nodes and landmark nodes in the output layer
in addition to an input node in the input layer. 
The nodes in the output layer are regularly located on a grid of the output space. 
They have location vectors that indicate the positions of the nodes in the output space, 
and codebook vectors to be trained. 
The codebook vector indicates a location in the input space.   
Only the landmark nodes are associated with landmark data, 
which is used for semi-supervised learning.
The codebook vectors are trained on 
data in the data-driven learning phase or 
landmark data in the landmark-driven phase. 
(2) {\it Relationship of data, landmark data, 
and codebook vectors in input space}. 
After completing the training of the LAMA, 
the codebook vectors cover the given data, 
retaining the relationships between the neighboring nodes in the output space. 
Note that the codebook vector of the landmark node (node 10) is 
the closest to the landmark data in the input space. 
A data point in the input space is projected onto a node in the output space, 
which has the closest codebook vector to the data in the input space.
Thus, the landmark nodes and the landmark data teach the LAMA the user-intended nonlinear projection.
}
\label{fig:concept}
\end{figure}

\begin{figure}[!h]
\includegraphics[width=\linewidth]{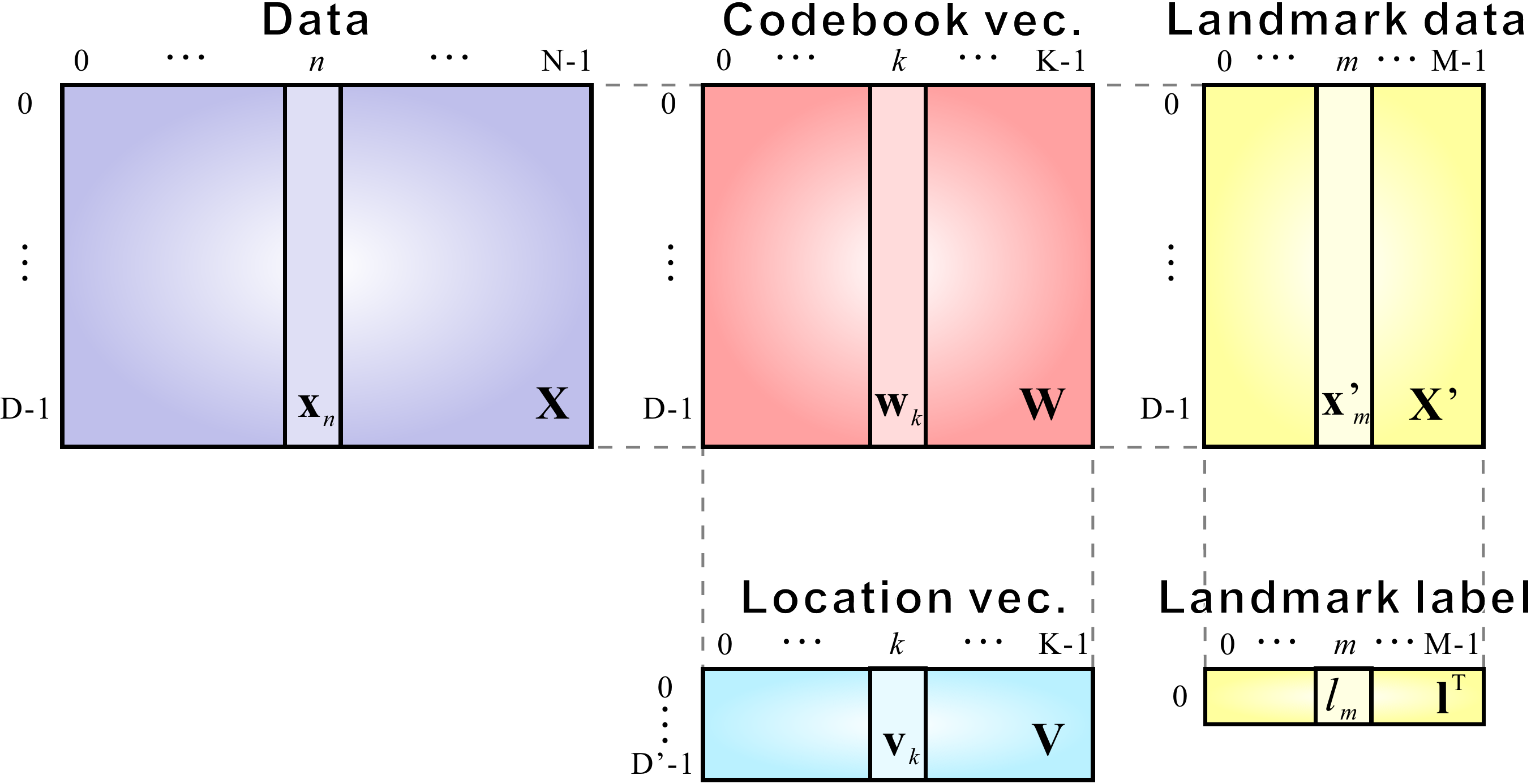}
\caption{Variables used in the LAMA. 
Data, landmark data, and codebook vectors are 
in the same dimension ($\mathrm{D}$-dimensional input space). 
The number of codebook vectors ($\mathrm{N}$) 
is the same as that of location vectors. 
Location vectors have different dimensions compared to codebook vectors ($\mathrm{D}'$). 
Landmark data are associated with the codebook vectors by scalar landmark labels.
}
\label{fig:variables}
\end{figure}

LAMA consists of numerous {\it standard nodes} and 
several landmark nodes in an output layer 
in addition to an input node in an input layer (see Fig.~\ref{fig:concept}(1)). 
The total number of nodes is $\mathrm{K}$.
The input node is connected to all the nodes in the output layer. 
Both standard and landmark nodes contain a {\it codebook vector} $\mathbf{w}_k \in  \Re^{\mathrm{D}}$ and a {\it location vector} 
$\mathbf{v}_k \in \Re ^{\mathrm{D}'}, k\in \left\lbrace 0,1,...,\mathrm{K}-1 \right\rbrace$ (see the size of the variables in Fig.~\ref{fig:variables}). 
Given $\mathrm{N}$ data points $\mathbf{x}_n \in \left\lbrace 0,1, ..., \mathrm{N}-1 \right\rbrace$, 
$\mathrm{M}$ landmark nodes that is associated with the {\it landmark data} $\mathbf{x}'_{m}  \in \left\lbrace 0,1, ..., \mathrm{M}-1 \right\rbrace$, 
and $\mathrm{M}$ {\it landmark labels} $l_m \in \left\lbrace 0,1,...,\mathrm{K}-1 \right\rbrace$,  
all the codebook vectors $\mathbf{w}_k, \forall k$ are trained. 
The combined data, codebook vectors, location vectors, landmark data, and landmark labels are denoted by 
$\mathbf{X} = (\mathbf{x}_0, ..., \mathbf{x}_{\mathrm{N}-1} )$, 
$\mathbf{W} = ( \mathbf{w}_0, ..., \mathbf{w}_{\mathrm{K}-1} )$, 
$\mathbf{V} =  ( \mathbf{v}_0, ..., \mathbf{v}_{\mathrm{K}-1} )$, 
$\mathbf{X'} = ( \mathbf{x'}_0, ..., \mathbf{x'}_{\mathrm{M}-1} )$, 
and $\mathbf{l} = ( l_0, ..., l_{\mathrm{M}-1} )^\mathrm{T}$, 
respectively, where $\cdot^\mathrm{T}$ indicates the transpose.

The locations of the standard and landmark nodes $\mathbf{V}$ are structurally arranged 
in a discrete $\mathrm{D}'$-dimensional output space (see Fig.~\ref{fig:concept}(1)). 
For example, 16 nodes are arranged in a $4 \times 4$ square architecture   
($\mathrm{K}=16, k \in \left\lbrace 0, 1, ..., \mathrm{K}-1 \right\rbrace, (\mathrm{K}_x \times \mathrm{K}_y)=(4 \times 4)$)
such that the distance from the neighboring nodes is 1 (no unit).  
In this example, node 15 has a location vector $\mathbf{v}_{15}=(3,3)^\mathrm{T}$.
Typically, the output space is two-dimensional, and the nodes 
have a square or hexagonal architecture, which enables us to 
easily visualize the U-matrix, which is a two-dimensional color contour plot \cite{Ultsch2003}. 
During learning the codebook vectors, the architecture is used, for example, to find neighboring nodes and to determine the update rate of each node.

The codebook vectors $\mathbf{W}$ can be represented in the input space together with data $\mathbf{X}$ and 
landmark data $\mathbf{X'}$  (see Fig.~\ref{fig:concept}(2)). 
After appropriate learning, the codebook vectors are fitted to a set of data (sometimes called manifold fitting), 
retaining the architecture of the nodes in the output space. 
For instance, the codebook vector of node 0 is located next to that of node 1. 
As shown in Fig.~\ref{fig:concept}(2), the red lines 
in the input space connect the codebook vectors of the nearest neighbor nodes in the output space.  
Thus, the codebook vectors are learned, 
retaining the relationships of the neighbor nodes in the output space.

An input data point is projected onto a node in the output space 
by finding the node that has the smallest Euclidean norm 
between the codebook vector and the input data in the input space. 
To realize the user-intended nonlinear projection, 
$\mathbf{W}$ is trained so that the codebook vector of the $m$-th landmark node $\mathbf{w}_{l_m}$ 
becomes the closest codebook vector to a corresponding landmark data point $\mathbf{x}'_m$ in the input space.  
In addition, the LAMA is updated by $\mathbf{X}$ so that $\mathbf{W}$ is fitted to the set of data retaining the architecture of $\mathbf{V}$. 
Such learning is realized by 
an {\it alternating update method} using data and landmark data.

\subsection{Alternating update method}
\begin{figure}[!h]
\includegraphics[width=6cm]{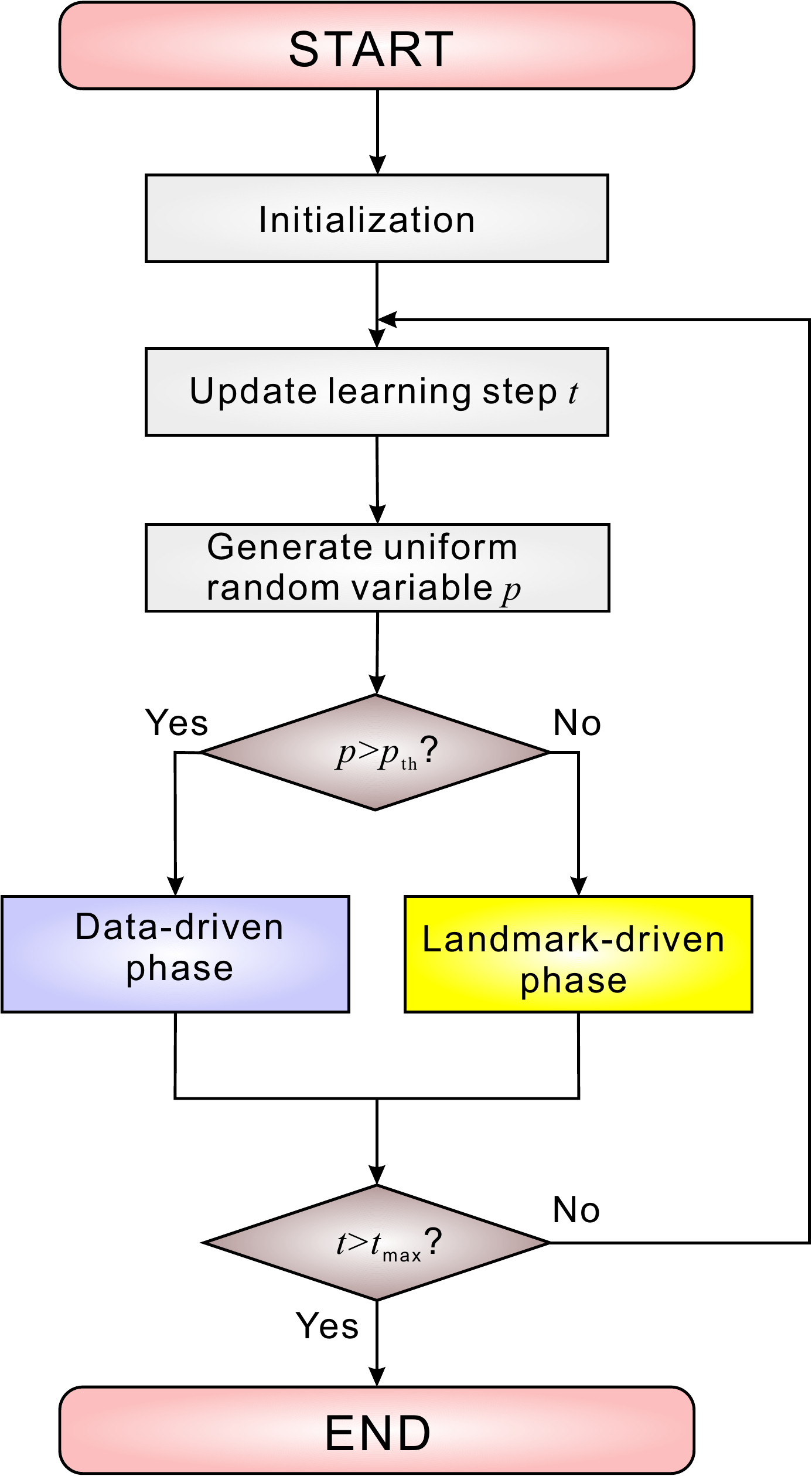}\\
\caption{Flowchart of learning phases. 
The LAMA is trained by the data-driven phase or the landmark-driven phase alternately.
One of the phases is selected in each learning step ($t$). 
If a randomly generated $p$ is greater than $p_\mathrm{th}$, 
the LAMA is trained by the data-driven phase (see Fig.~\ref{fig:phases}(1)). 
If $p$ is less than $p_\mathrm{th}$, the LAMA is updated by the 
landmark-driven phase (see Fig.~\ref{fig:phases}(2)). 
This process is repeated $t_\mathrm{max}$ times. 
}
\label{fig:flowchart}
\end{figure}

\noindent
Figure \ref{fig:flowchart} shows the flowchart of training the LAMA. 
First, all the variables and parameters are initialized. 
In this study, $\mathbf{W}$ is initialized by a uniformly distributed random number ranging from 0 to 1 
in order to observe the robustness of the initial values.
Further, $\mathbf{W}$ is iteratively updated in each discrete 
learning step $t \in \left\lbrace 0,1, ..., t_\mathrm{max}-1 \right\rbrace$. 
In each $t$, {\it data-driven} or {\it landmark-driven phases} are alternately selected 
in the ratio $(1-p_\mathrm{th}):p_\mathrm{th}$, where $0 \leq p_\mathrm{th}<1$. 
However, a user-intended projection designated by the landmark nodes 
cannot be obtained by simply repeating the above-mentioned procedure, 
due to the complexity of the parameters. 
Different learning functions of the two phases seem to be useful for obtaining the projection.

\subsubsection{Data-driven phase}

\begin{figure}[!h]
\includegraphics[width=\linewidth]{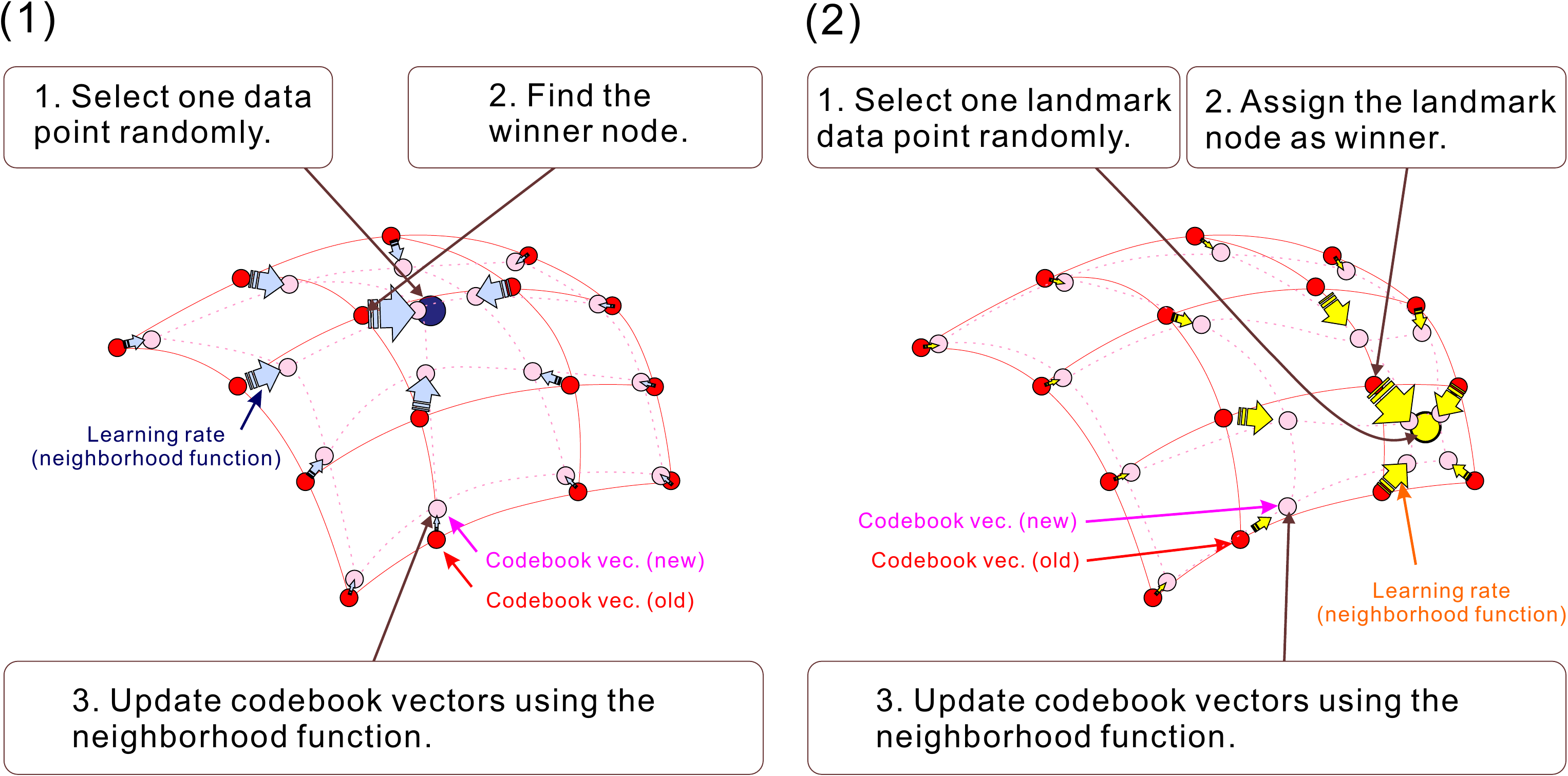}
\caption{Alternating update method. 
(1) {\it Data-driven phase}. When the data-driven phase is activated, one data point from the given data is selected randomly. 
Next, the winner node, i.e., the node having the codebook vector 
that is the nearest to the data point, is chosen. 
Finally, all the codebook vectors are updated using the neighborhood function. 
(2) {\it Landmark-driven phase}. When the landmark-driven phase is performed, 
a landmark data point is randomly selected. 
Next, the node having the landmark data point 
is assigned as the winner node. 
Finally, all the codebook vectors are updated by the neighborhood function.  
}
\label{fig:phases}
\end{figure}

\noindent
The data-driven phase updates the codebook vectors such that they fit the entire dataset (see Fig.~\ref{fig:phases}(1)).  
In the data-driven phase, a data point $n$ is randomly selected as the input of the LAMA. 
Then, the node that has the closest codebook vector to the input data is selected as the winner node. 
Thus, the winner node $k_\mathrm{d}$ is found such that it minimizes 
the Euclidean norm between $\mathbf{x}_n$ and $\mathbf{w}_k$: 
\begin{eqnarray}
k_\mathrm{d}(n)=\argmin_k \left\Vert \mathbf{x}_n - \mathbf{w}_k \right\Vert^2,
\label{eq:Data_driven_BMU}
\end{eqnarray}
where $\left\Vert \cdot \right\Vert$ represents the Euclidean norm. For simplicity, the node 
$k_\mathrm{d}(n)$ is also denoted as $k_\mathrm{d}$. 
Next, $\mathbf{W}$ is updated by Hebbian learning 
considering the distance from the winner node to a focused node $k$ in the output space 
in addition to the learning step. 
Such learning is implemented by a learning rate function called the {\it neighborhood function}, 
denoted by $\alpha_k(t)$:
\begin{eqnarray}
 \alpha_k (t)= a(t) \cdot h_\mathrm{a}(k_\mathrm{d},k,\sigma(t)), \forall k.
\end{eqnarray} 
\noindent
Here, $a(t)$ is an exponentially descending function of $t$ that determines the maximum learning rate: 
\begin{eqnarray}
a(t)=a_\mathrm{min} + (a_\mathrm{max} - a_\mathrm{min}) \exp \left(-\frac{t}{\tau_a} \right),
\end{eqnarray}
where $a_\mathrm{max}$, $a_\mathrm{min}$, and $\tau_a$ denote the maximum, minimum, and the time decay of the learning rate, respectively; 
and $h_\mathrm{a}(k_\mathrm{d},k,\sigma(t))$ determines the spread of the learning rate by the Gaussian distribution, 
considering the distance between node $k$ and winner node $k_\mathrm{d}$: 
\begin{eqnarray}
h_\mathrm{a}(k_\mathrm{d},k,\sigma(t)) = \exp \left(- \frac{\left\Vert \mathbf{v}_{k_\mathrm{d}} - \mathbf{v}_{k} \right\Vert^2}{2\sigma^2(t)}  \right). 
\end{eqnarray}
Here, $\sigma (t)$ is the exponentially descending function of $t$ that determines the spread of the above-mentioned Gaussian distribution: 
\begin{eqnarray}
\sigma(t)=\sigma_\mathrm{min} + (\sigma_\mathrm{max} - \sigma_\mathrm{min}) \exp \left(-\frac{t}{\tau_\sigma}\right),
\end{eqnarray}
where $\sigma_\mathrm{max}$, $\sigma_\mathrm{min}$, and $\tau_\sigma$ denote the maximum, minimum, and time decay of the extent, respectively. 
Therefore, $\alpha_k (t)$ decays as the distance from the winner node $k_\mathrm{d}$ to node $k$
($\left\Vert \mathbf{v}_{k_\mathrm{d}} - \mathbf{v}_{k} \right\Vert^2$) 
increases, and as $t$ increases. 
Finally, $\mathbf{W}$ is updated as follows:
\begin{eqnarray}
\mathbf{w}_k  \leftarrow \mathbf{w}_k + \alpha_k (t) \cdot \left( \mathbf{x}_n - \mathbf{w}_k \right), \forall k.
\end{eqnarray}
During the data-driven phase, the landmark node behaves in the same way as the standard node. 
This implies that the landmark node can be a winner node and 
its codebook vector is updated in this phase 
without using the landmark data or landmark labels.

\subsubsection{Landmark-driven phase}
\begin{figure}[!h]
\includegraphics[width=\linewidth]{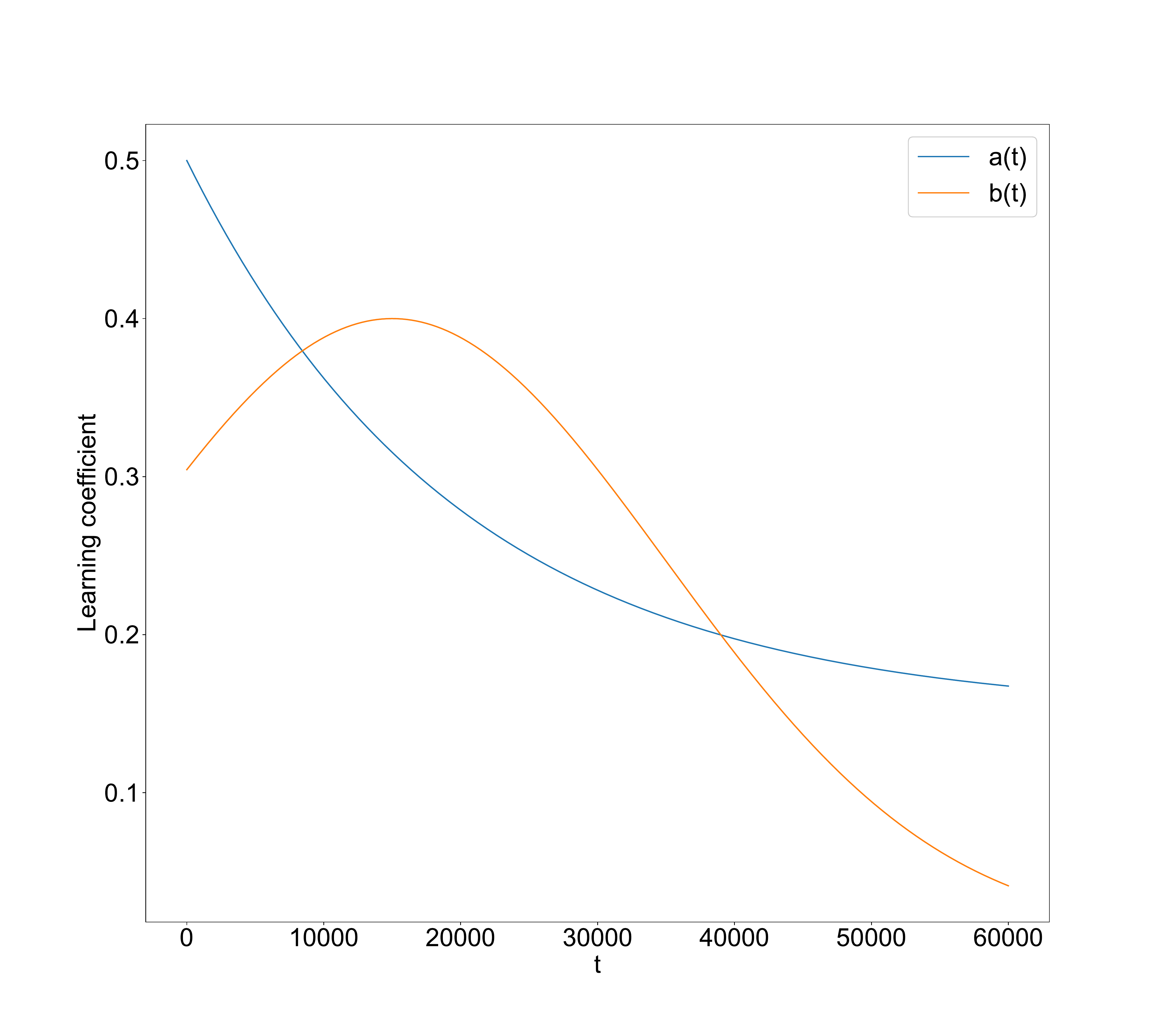}
\caption{Example of $a(t)$ and $b(t)$. 
Maximum learning rate of the neighborhood function is defined 
by $a(t)$ and $b(t)$ in the data-driven and landmark-driven phases, respectively, 
where $a(t)$ is a monotonically descending exponential function 
and $b(t)$ is a Gaussian function having a peak in the middle of the learning steps. 
}
\label{fig:paramExampleAB}
\end{figure}

\begin{figure}[!h]
\includegraphics[width=\linewidth]{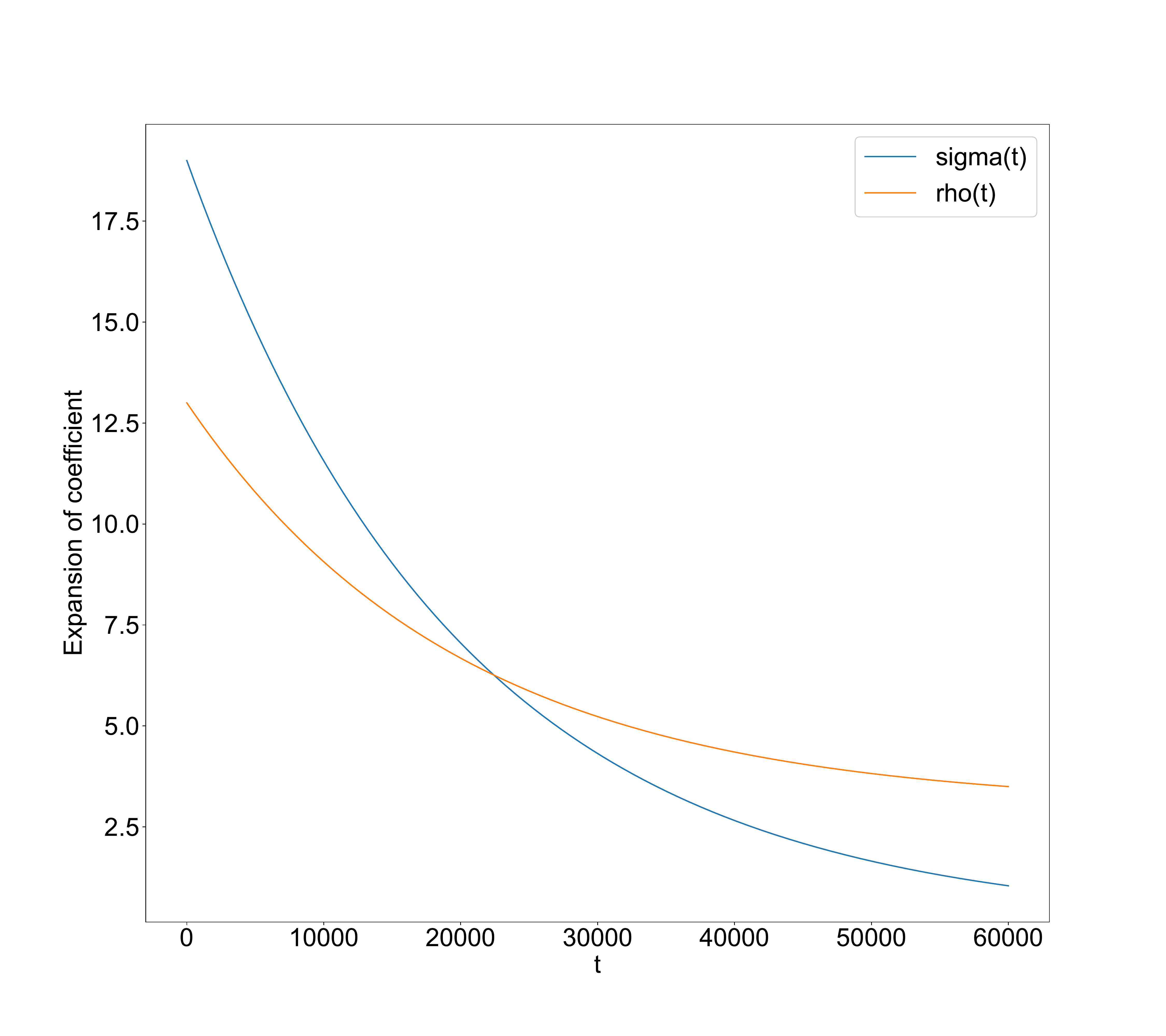}
\caption{Example of $\sigma(t)$ and $\rho(t)$. 
The stretch of the learning rate in the neighborhood function 
is represented by $\sigma(t)$ and $\rho(t)$ 
in the data-driven and landmark-driven phases, respectively. 
They are monotonically descending exponential functions with different slopes. 
}
\label{fig:paramExampleSigRho}
\end{figure}

\noindent
On the other hand, the landmark-driven phase makes 
the codebook vector of the landmark node close to its landmark data 
(see Fig.~\ref{fig:phases}(2)). 
In the landmark-driven phase, a landmark data point $m$ is randomly selected as the input of the LAMA. 
Next, the corresponding landmark node is assigned to a winner node $(k_\mathrm{l}(m)=l_m)$ 
regardless of the distance from the landmark data point. 
As with the data-driven phase, the neighborhood function $\beta_k(t)$ is used, which is independent of $\alpha_k(t)$:
\begin{eqnarray}
 \beta_k (t)= b(t) \cdot h_\mathrm{b}(k_\mathrm{l},k,\rho(t)), \forall k.
\end{eqnarray} 
Here, $b(t)$ is a Gaussian function of $t$ that determines the maximum learning rate: 
\begin{eqnarray}
b(t)=b_\mathrm{min} + (b_\mathrm{max} - b_\mathrm{min}) \exp \left(-\frac{\left\Vert t - t_\mathrm{center} \right\Vert^2}{2 \rho_b^2} \right),
\end{eqnarray}
where $b_\mathrm{max}$ and $b_\mathrm{min}$ denote the maximum and minimum of the learning rate, respectively; 
$t_\mathrm{center}$ is the step that has the maximum learning rate; 
and $\tau_b$ indicates the distribution of the above-mentioned Gaussian function over the steps.  
Further, $h_\mathrm{b}(k_\mathrm{l},k,\rho(t))$ assigns the spread of the learning rate by the Gaussian distribution: 
\begin{eqnarray}
h_\mathrm{b}(k_\mathrm{l},k,\rho(t)) = \exp \left(- \frac{\left\Vert \mathbf{v}_{k_\mathrm{l}} - \mathbf{v}_{k} \right\Vert^2}{2\rho^2(t)}  \right). 
\end{eqnarray}
Here, $\rho (t)$ is the exponentially descending function of $t$ that determines the extent of the above-mentioned Gaussian distribution:
\begin{eqnarray}
\rho(t)=\rho_\mathrm{min} + (\rho_\mathrm{max} - \rho_\mathrm{min}) \exp \left(-\frac{t}{\tau_\rho}\right),
\end{eqnarray}
where $\rho_\mathrm{max}$, $\rho_\mathrm{min}$, and $\tau_\rho$ denote the maximum, minimum, and time decay of the extent, respectively. 
Therefore, $\beta_k (t)$ decays as the distance from the winner node $k_\mathrm{l}$ to node $k$ increases, and as $t$ increases. 
Finally, $\mathbf{W}$ is updated as follows:
\begin{eqnarray}
\mathbf{w}_k  \leftarrow \mathbf{w}_k + \beta_k (t) \cdot \left( \mathbf{x}'_m - \mathbf{w}_k \right), \forall k.
\end{eqnarray}
The landmark-driven and data-driven phases 
have different neighborhood functions that cope with the two types of learning. 
An example of their parameters is shown in Fig.~\ref{fig:paramExampleAB}) and Fig.~\ref{fig:paramExampleSigRho}). 
When $t=0$, $a(t)$ is stronger than $b(t)$ (see the example in Fig.~\ref{fig:paramExampleAB}). 
When $t=t_\mathrm{center}=1500$, $b(t)$ is larger than $a(t)$.
Then, they decay monotonically. 
In the end ($t=t_\mathrm{max}-1=59999$), $a(t)$ is stronger than $b(t)$ again. 
Further, $\sigma(t)$ and $\rho(t)$ decay monotonically with different parameters 
(see the example in Fig.~\ref{fig:paramExampleSigRho}). The function 
$\sigma(t)$ is stronger than $\rho(t)$ when $t=0$; 
however, $\sigma(t)$ decays faster.

\subsection{Visualization method}
\noindent
The learning properties of SOM and LAMA were compared 
by visualizing their codebook vectors and data in the input space.  
As the codebook vectors and data are typically in high-dimensional input space, 
the dimension was reduced to two or three for the visualization. 
Principal component analysis (PCA) was adopted, 
and the first three principal components were used 
for the visualization. 
The neighboring codebook vectors were connected by lines 
to clarify the relationship between the neighbors.

In addition, the data labels were projected onto a two-dimensional output space 
and then visualized by the U-matrix \cite{Ultsch2003}.
This study used a simplified U-matrix defined as follows. 
First, a $\mathrm{K}_x\times \mathrm{K}_y$ matrix 
$\mathbf{U} \in \Re^{\mathrm{K}_x \times \mathrm{K}_y}$ 
was initialized. 
Each component of $\mathbf{U}$ was associated with the nodes in the output space. 
Focusing on a node $k$ located at $(k_x, k_y)$ in the output space, 
four neighbors of the node, that is, nodes that have the minimum distance from the node $k$, were selected. 
Then, the summation of the Euclidean distance of the codebook vectors between node $k$ and the neighboring nodes
was calculated.
Assuming that the number of neighboring nodes is $\mathrm{N}_n$
and the neighbor nodes are indexed as $k_n, n\in \left\lbrace 1,2,..., \mathrm{N}_n\right\rbrace$, 
the component of the U-matrix at $(k_x, k_y)$ is
\begin{eqnarray}
u_{k_x, k_y}=\sum_{k_n=1}^{\mathrm{N}_n} \left\Vert \mathbf{w}_k - \mathbf{w}_{k_n} \right\Vert^2.
\end{eqnarray}
This simplified U-matrix can be visualized as a contour plot. 
The distance between the neighbors is visualized in color.

\subsection{Numerical evaluation of error indices}
\noindent
To evaluate the goodness of the codebook vectors of the SOM numerically, 
quantization and topographic errors (TE) have been applied \cite{Meschino2015}. 
However, they do not always explain the error of LAMA 
because they do not take landmarks into consideration. 
This study employed extended error indices to clarify 
the learning properties of LAMA and SOM.

\subsubsection{Quantization Error of Data}
\noindent
The quantization error of data (QED) represents the mean distance   
between each data point and the corresponding winner node in input space. 
The QED can be calculated as follows:
\begin{eqnarray}
E_\mathrm{QED}=\frac{1}{\mathrm{N}} \sum_{n=1} ^{\mathrm{N}}
\left\Vert \mathbf{x}_n - \mathbf{w}_{k_{\mathrm{d}}(n)} \right\Vert,
\end{eqnarray}
where the winner node of the $n$-th data point is denoted by $k_{\mathrm{d}}(n)$ 
(see Eq.~\ref{eq:Data_driven_BMU}). 
If the QED is small, the codebook vectors 
are successfully distributed over the data.
The QED should be increased if 
part of the data is far from any codebook vectors.

\subsubsection{Quantization Error of Landmark}
\noindent
The QED represents errors between data and codebook vectors. 
On the other hand, the quantization error of landmark (QEL)
indicates errors between landmark data and corresponding codebook vectors.
It can be estimated as follows: 
\begin{eqnarray}
E_\mathrm{QEL}=\frac{1}{\mathrm{M}} \sum_{m=1} ^{\mathrm{M}}
\left\Vert \mathbf{x'}_m - \mathbf{w}_{k_{\mathrm{l}}(m)} \right\Vert,
\end{eqnarray}
where the winner node of the $m$-th landmark data point is denoted by 
$k_{\mathrm{l}}(m)$. 
The QEL should be small if landmark data and corresponding landmark nodes 
are closely located in the input space.

\subsubsection{Square Topographic Error}
\noindent
The TE verifies how well the topology is preserved. 
In other words, the first-place winner node of the $n$-th data point $k_{\mathrm{d}(n)}$, and 
the second-place winner node of the $n$-th data point $k_{\mathrm{d}(n)}'$ 
should be adjacent to each other. 
It can be computed by a function that judges 
whether the nodes of $k_{\mathrm{d}}$ and $k_{\mathrm{d}}'$ are 
larger than the minimum distance of nodes: 
\begin{eqnarray}
s_{\mathrm{TE}}(n)=\left\lbrace
 \begin{array}{cc}
 1: & if \left\Vert \mathbf{v}_{k_{\mathrm{d}}'(n)}- \mathbf{v}_{k_{\mathrm{d}}(n)} \right\Vert > \mathrm{D_{TE}} + \epsilon.   \\ 
 0: & otherwise.
 \end{array} 
\right.
\end{eqnarray}
where the minimum distance to the closest node in the output space 
is denoted by $\mathrm{D_{TE}}$; and $\epsilon=0.01$ indicates a margin.
In this study, $\mathrm{D_{TE}}=1$.
Using that function, the TE can be estimated as follows:
\begin{eqnarray}
E_{\mathrm{TE}}=\frac{1}{\mathrm{N}} \sum _{n=1} ^{\mathrm{N}}  s_{\mathrm{TE}}(n).
\end{eqnarray}
The lower the TE becomes, the more the topology is preserved.

In this study, the TE is extended to consider eight neighbors of each node using square topology. 
The extended error index is named the square topographic error (STE). 
Since the maximum distance of eight neighbors is $\mathrm{D_{STE}}=\sqrt{2} \times {\mathrm{D_{TE}}}$, 
$s_{\mathrm{TE}}(n)$ was modified to consider $\mathrm{D_{STE}}$ as follows:
\begin{eqnarray}
s_{\mathrm{STE}}(n)=\left\lbrace
 \begin{array}{cc}
 1: & if \left\Vert \mathbf{v}_{k_{\mathrm{d}}'(n)}- \mathbf{v}_{k_{\mathrm{d}}(n)} \right\Vert > \mathrm{D_{STE}} + \epsilon.  \\ 
 0: & otherwise.
 \end{array} 
\right.
\end{eqnarray}
Instead of TE, the following STE was employed: 
\begin{eqnarray}
E_{\mathrm{STE}}=\frac{1}{\mathrm{N}} \sum _{n=1} ^{\mathrm{N}}  s_{\mathrm{STE}}(n).
\end{eqnarray}
It becomes large if the second-place winner node of 
a focused node is selected out of eight neighbors.

\section{Zoo dataset analysis}

\subsection{Dataset and parameters}
\noindent
To reveal the differences between the learning properties of LAMA and SOM, 
they were applied to the 
Zoo dataset of the UCI Machine Learning Repository (
\url{https://archive.ics.uci.edu/ml/datasets/zoo}). 
The dataset contains 101 animal names with 17 attributes, such as 
whether they have hair, feathers, and lay eggs. 
Note that it contains two instances of ``frog'' and one instance of ``girl.'' 
The attributes are described using Boolean (0 or 1) or numeric (e.g., integer) form. 
In this study, 16 attributions, with the exception of the type, were used. 
This dataset was selected because a similar animal dataset was used for 
the benchmark of the SOM \cite{kohonen2001}. 
In addition, it may explains how to represent features of, 
for example, food, clothes, and other products in the market.

The learning properties of LAMA and SOM were assessed 
by visualizing the structure of the codebook vectors in the input space. 
Furthermore, the location of the data and the landmark data were displayed in a U-matrix.
The detailed experimental parameters used for learning are presented in Table~\ref{table:paramZoo}. 
To demonstrate the learning properties of LAMA, 
four different examples of landmarks were applied. 
In LAMA1, a sea lion ($n=75$) was associated with the center node of the output space ($k=312$). 
In LAMA2, a duck ($n=21$) and a penguin ($n=58$) were assigned 
to a middle left node ($k=303$) and a middle right node ($k=321$), respectively.  
Similarly, LAMA3 contained three landmark nodes that showed 
the relationships between a mink ($n=48$) and a top center node ($k=37$), 
between a seal ($n=74$) and a bottom left node ($k=552$), 
and between a slowworm ($n=80$) and a bottom right node ($k=572$). 
Finally, LAMA4 was trained with landmark nodes located at four edges ($k=0,24,600,624$), 
which were paired with the mink ($n=48$), the toad ($n=89$), the seal ($n=74$), and the slowworm ($n=80$), respectively.

\begin{table}[!h]
\centering
\caption{
{Parameters of SOM and LAMA used for the Zoo dataset analysis.}}
\begin{tabular}{c|ccccc}
  	\hline
    Algorithm 			& SOM 			& LAMA1 		& LAMA2 		& LAMA3 		& LAMA4\\
    \hline
    $\mathrm{K}_x$ 				& 25 			& 25 			& 25 			& 25			& 25  \\
    $\mathrm{K}_y$ 				& 25 			& 25 			& 25 			& 25			& 25  \\
    $t_\mathrm{max}$ 			& 60000 		& 60000 		& 60000 		& 60000 		& 60000 \\
    $a_\mathrm{max}$ 			& 0.5 			& 0.5 			& 0.5 			& 0.5 			& 0.5  \\
    $a_\mathrm{min}$ 			& 0.15 			& 0.15 			& 0.15 			& 0.15 			& 0.15 \\
    $\tau_a$			& $t_\mathrm{max}/3-1$	& $t_\mathrm{max}/3-1$	& $t_\mathrm{max}/3-1$	& $t_\mathrm{max}/3-1$	& $t_\mathrm{max}/3-1$ \\
    $\sigma_\mathrm{max}$		& 19			& 19			& 19			& 19			& 19 \\
    $\sigma_\mathrm{min}$		& 0.1			& 0.1			& 0.01			& 0.01			& 0.01 \\
    $\tau_\sigma$		& $t_\mathrm{max}/3-1$	& $t_\mathrm{max}/3-1$	& $t_\mathrm{max}/3-1$	& $t_\mathrm{max}/3-1$	& $t_\mathrm{max}/3-1$ \\
    $b_\mathrm{max}$ 			& --			& 0.4 			& 0.4 			& 0.4 			& 0.4  \\
    $b_\mathrm{min}$ 			& -- 			& 0.01 			& 0.075 		& 0.075 		& 0.1 \\
    $\tau_b$			& --			& $t_\mathrm{max}/3-1$	& $t_\mathrm{max}/3-1$	& $t_\mathrm{max}/3-1$	& $t_\mathrm{max}/3-1$ \\
    $t_\mathrm{center}$ 		& --	 		& 15000 		& 15000 		& 15000 		& 15000  \\
    $\rho_{b}$			& --			& 20000			& 25000			& 25000			& 25000	 \\
    $\rho_\mathrm{max}$		& --			& 13			& 13			& 13			& 13			\\
    $\rho_\mathrm{min}$		& --			& 3				& 0.7			& 1				& 1.5 \\
    $\tau_\rho$			& --			& $t_\mathrm{max}/3-1$	& $t_\mathrm{max}/3-1$	& $t_\mathrm{max}/3-1$	& $t_\mathrm{max}/3-1$ \\
    $p_\mathrm{th}$			& --			& 0.01			& 0.05			& 0.07			& 0.09 \\
    Landmark 1 $(n:k)$	& --			& 75:312		& 21:303		& 48:37			& 48:0 \\
    Landmark 2 $(n:k)$	& --			& --			& 58:321		& 74:552		& 89:24 \\
    Landmark 3 $(n:k)$	& --			& --			& --			& 80:572		& 74:600 \\
    Landmark 4 $(n:k)$	& --			& --			& --			& --			& 80:624 \\
    \hline
  \end{tabular}
\begin{flushleft} 
\end{flushleft}
\label{table:paramZoo}
\end{table}

In addition, numerical learning properties of SOM and LAMA 
were assessed using QED, QEL, and STE. 
SOM, LAMA1, LAMA2, LAMA3, and LAMA4 were trained 
100 times each. 
Next, the mean values of QED, QEL, and STE   for each algorithm 
were visualized in seven learning iterations 
(t = 0,9999, 19999, 29999, 39999, 49999, 59999).

\subsection{Visualization of learning properties}
\begin{figure}[!h]
\includegraphics[width=\linewidth]{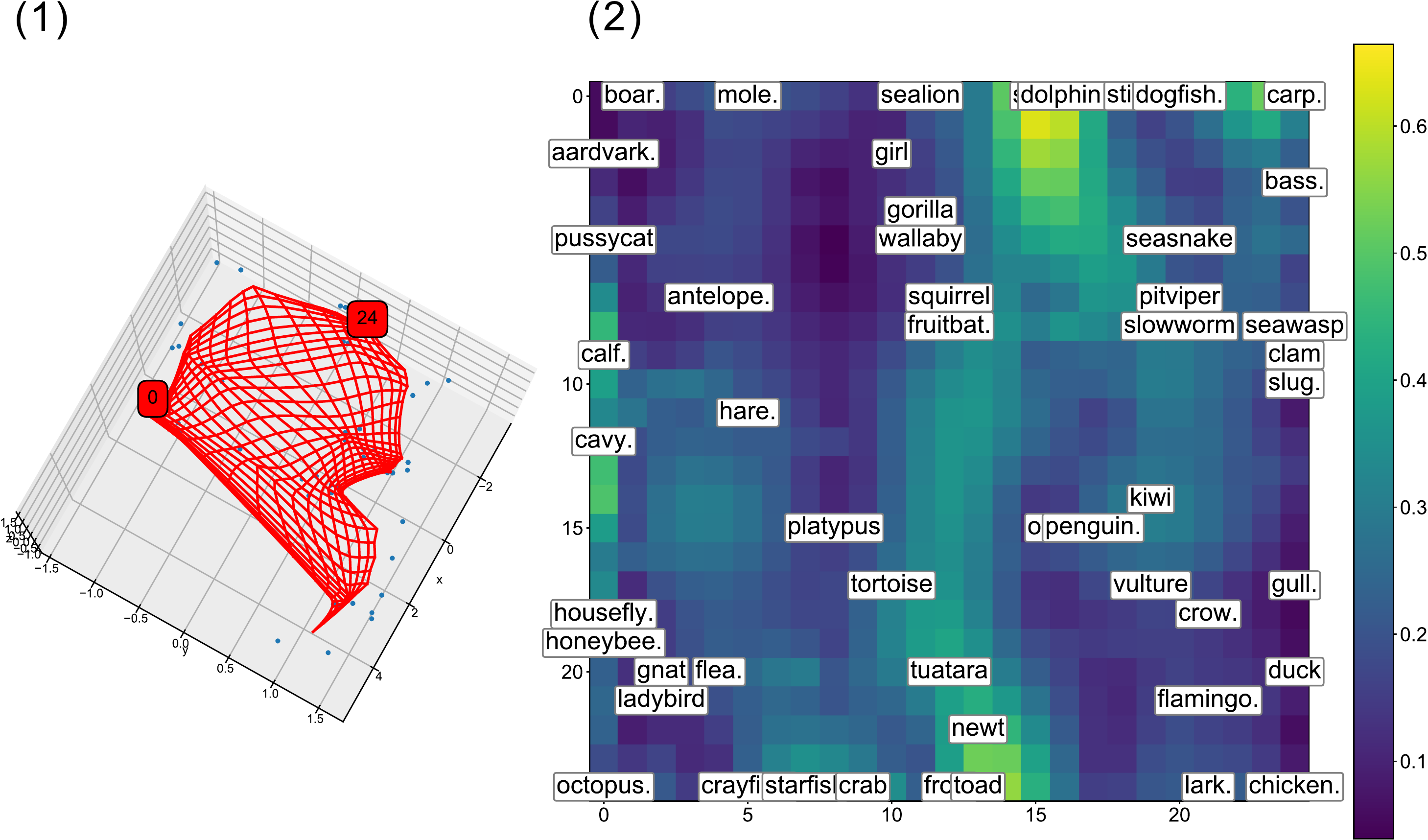}
\caption{Codebook vectors and U-matrix of the SOM trained from the Zoo dataset.
(1) {\it Codebook vectors and data in the output space}. 
Nodes 0 and 24 are displayed by red labels
to illustrate the rotation of the codebook vectors.
(2) {\it Data labels in the U-matrix}. 
The white labels indicate the data labels 
projected onto the location of the U-matrix. 
The dot in the label implies that multiple labels are projected onto the same node. 
The color of the U-matrix represents the distance from the neighboring nodes. 
As the color turns yellow, the distance increases. }
\label{fig:zooSom}
\end{figure}

\noindent
For comparison with LAMA, SOM was trained on the Zoo dataset, 
and the codebook vectors and the U-matrix were then visualized. 
Fig.~\ref{fig:zooSom}(1) shows the codebook vectors and the data points. 
The red mesh grid composed of lines between the neighboring codebook vectors cover the data smoothly, 
retaining the architecture of the nodes in the output space. 
The density of the codebook vectors seems to be similar in every part of the data. 
Fig.~\ref{fig:zooSom}(2) represents the U-matrix of the SOM. 
The color of the U-matrix represents the summation of the distance 
in the input space between four immediate neighbor nodes, 
where the neighbor nodes were determined by the distance in the output space. 
The blue part of the U-matrix indicates that the neighboring codebook vectors were close, 
while the yellow part indicates that the neighboring codebook vectors were far.
Each animal data point was projected onto the U-matrix (Fig.~\ref{fig:zooSom}(1)). 
Animals with similar features, such as the platypus and the tortoise, 
were closely located in the U-matrix. 
On the other hand, animals with different features, such as the gorilla and the crow, 
were far in the U-matrix, 
separated by a yellow boundary.

\begin{figure}[!h]
\includegraphics[width=\linewidth]{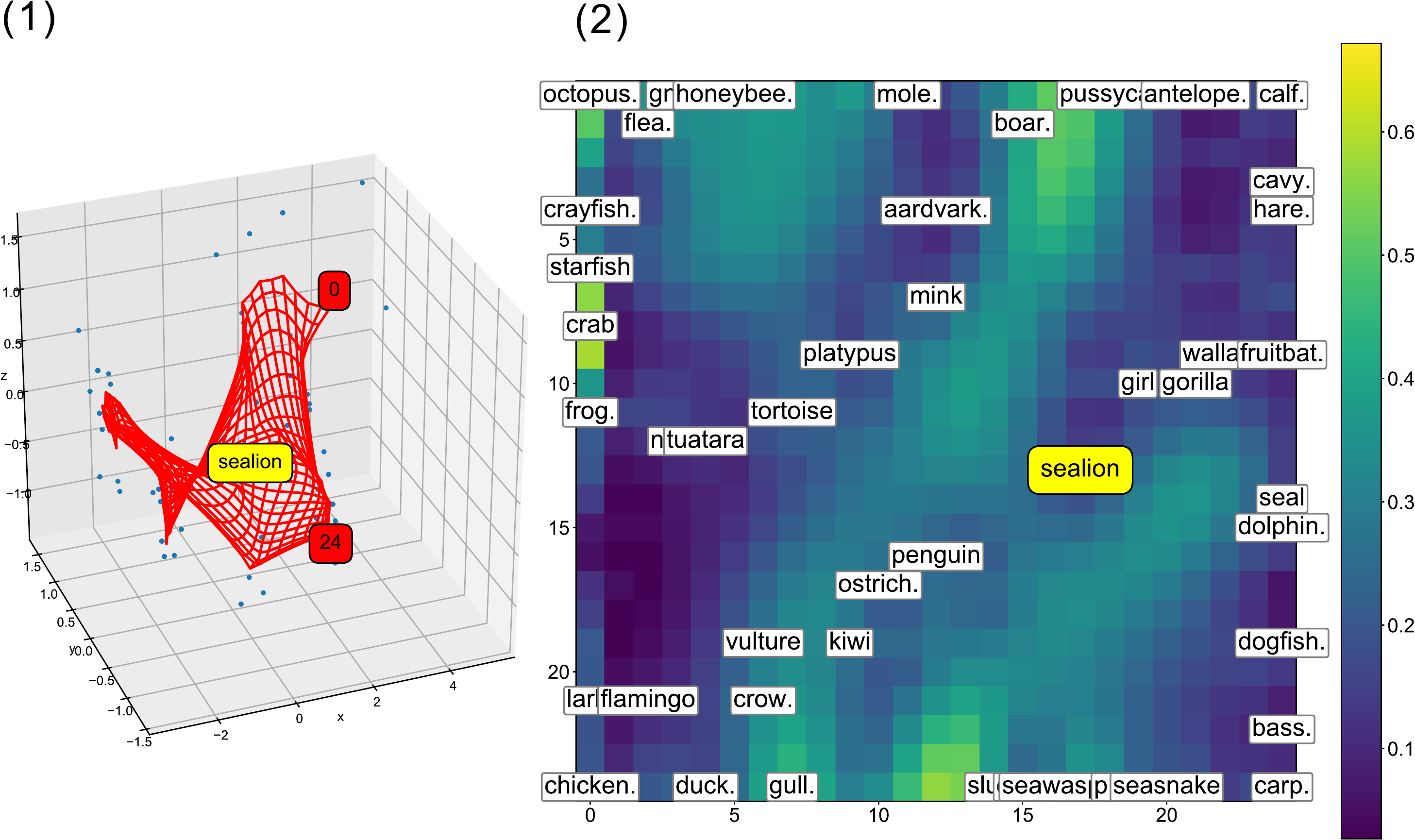}
\caption{Codebook vectors and U-matrix of LAMA1 trained using the Zoo dataset. 
(1) {\it Codebook vectors and data in the output space}. 
Nodes 0 and 24 are shown by red labels
to illustrate the rotation of the codebook vectors.
The yellow label indicates the location of the landmark data. 
(2) {\it Data labels on the U-matrix}. 
The white labels indicate data projected onto the location of the U-matrix. 
The yellow label indicates the landmark data. 
The dot in the label implies that multiple labels are projected onto the same node. 
The color of the U-matrix represents the distance from the neighboring nodes. 
As the color turns yellow, the distance increases. 
}
\label{fig:zooLama1}
\end{figure}

LAMA1 provided a different view of the data. 
The codebook vectors of LAMA1 are shown in Fig.~\ref{fig:zooLama1}(1). 
The red mesh grid covered the data 
with its center close to the landmark data shown in yellow label. 
Fig.~\ref{fig:zooLama1}(2) shows the U-matrix of LAMA1. 
The landmark data shown in yellow were approximately centered. 
In the U-matrix of the SOM (Fig.~\ref{fig:zooSom}(2)), 
the label was located near an edge. 
In addition, the data around the landmark data (sea lion) in the U-matrix of LAMA1 were widely distributed, 
while those of the SOM were not. 
Thus, LAMA1 in this example provided landmark-centered visualization.

\begin{figure}[!h]
\includegraphics[width=\linewidth]{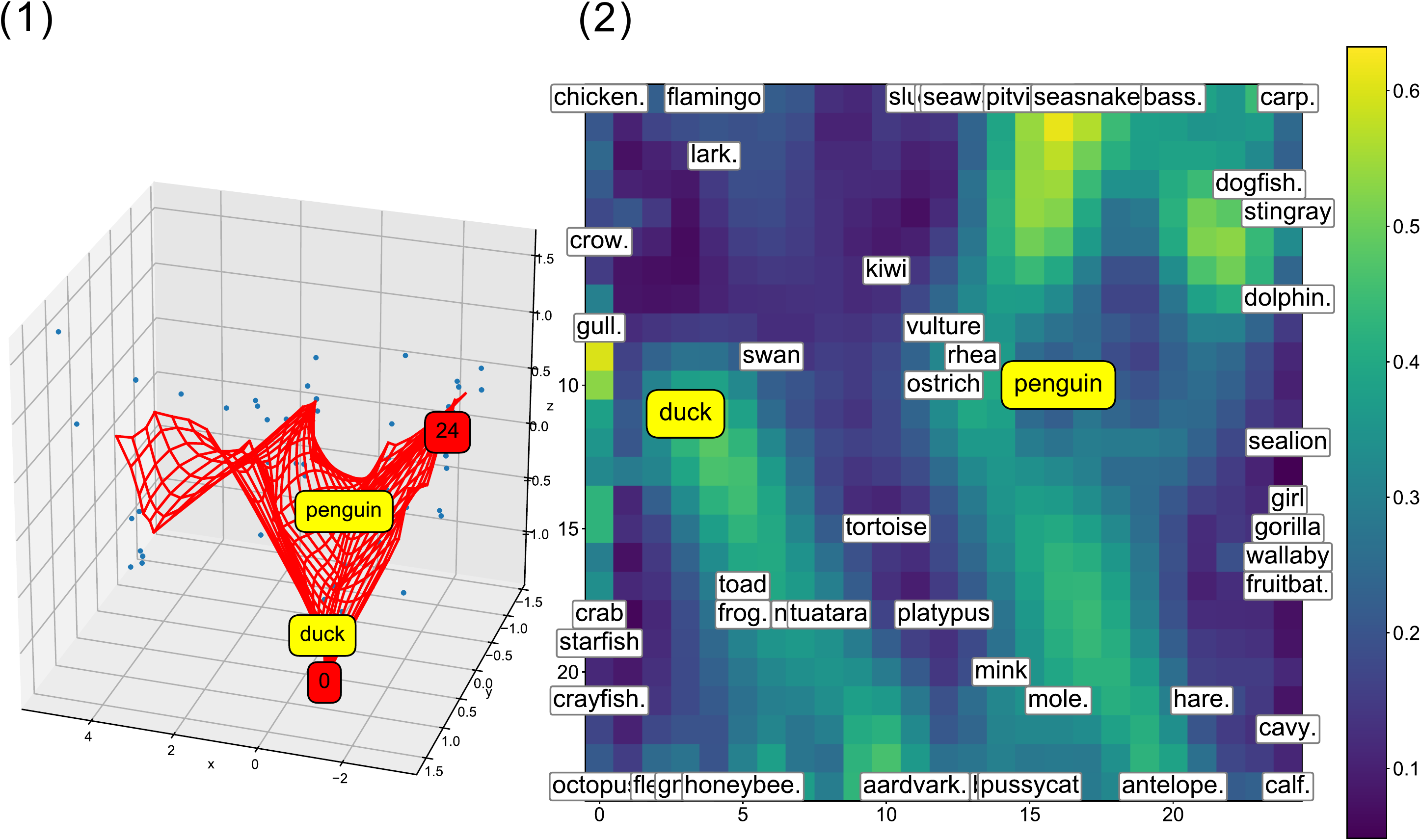}
\caption{Codebook vectors and the U-matrix of LAMA2 trained using the Zoo dataset. 
(1) {\it Codebook vectors and data in the output space}. 
Nodes 0 and 24 are displayed by red labels
to illustrate the rotation of the codebook vectors.
The yellow labels indicate the location of the landmark data. 
(2) {\it Data labels on the U-matrix}. 
The white labels indicate data projected onto the location of the U-matrix. 
The yellow labels indicate the landmark data.
The dot in the label implies that multiple labels are projected onto the same node. 
The color of the U-matrix represents the distance from the neighboring nodes. 
As the color turns yellow, the distance increases. 
}
\label{fig:zooLama2}
\end{figure}

LAMA2 shows the relationship of the data around two designated landmarks. 
The codebook vectors of LAMA2 are shown in Fig.~\ref{fig:zooLama2}(1). 
The mesh grid was shifted so that the two landmark data points
were close to the center line of the grid. 
The density of the codebook vectors did not appear uniform, 
which was different from the mesh grid of the SOM (see Fig.~\ref{fig:zooSom}(1)). 
Fig~\ref{fig:zooLama2}(2) shows the U-matrix of LAMA2. 
The two landmark data points were projected onto the middle left and the middle right, respectively. 
The data points between the two landmark data points were clearly seen as if they were zoomed in, 
compared with the results of SOM and LAMA1. 
As seen above, LAMA2 provided enlarged data visualization focusing on two landmark data points.

\begin{figure}[!h]
\includegraphics[width=\linewidth]{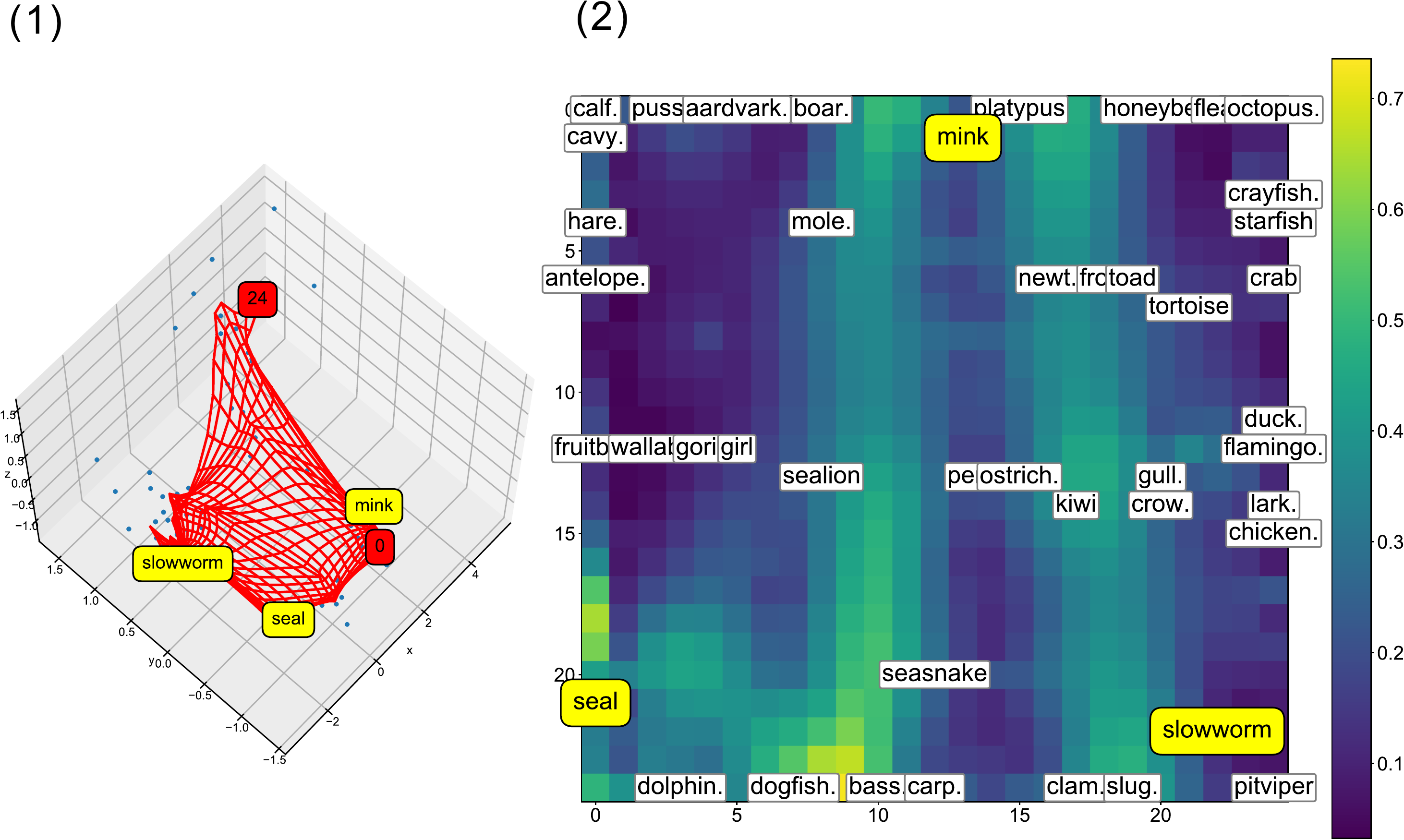}
\caption{Codebook vectors and the U-matrix of LAMA3 trained using the Zoo dataset. 
(1) {\it Codebook vectors and data in the output space}. 
Nodes 0 and 24 are shown by red labels
to illustrate the rotation of the codebook vectors.
The yellow labels indicate the location of landmark data. 
(2) {\it Data labels on the U-matrix}. 
The white labels indicate data projected onto the location of the U-matrix. 
The yellow labels indicate the landmark data.
The dot in the label implies that multiple labels are projected onto the same node. 
The color of the U-matrix represents the distance from the neighboring nodes. 
As the color turns yellow, the distance increases. 
}
\label{fig:zooLama3}
\end{figure}

Similar to LAMA2, LAMA3 showed the relationship of the data surrounded by three landmark data points. 
Fig~\ref{fig:zooLama3}(1) represents the codebook vectors of LAMA3. 
Three different landmark data points, namely, the mink, the seal, and the slowworm, 
were located close to the top center, bottom left, and bottom right of the mesh grid, 
respectively. 
Fig~\ref{fig:zooLama3}(2) shows the U-matrix of LAMA3. 
The data points encircled by these landmarks were widely displayed on the U-matrix. 
As the three landmark data points had different features, 
they were divided by the yellow boundary on the U-matrix.

\begin{figure}[!h]
\includegraphics[width=\linewidth]{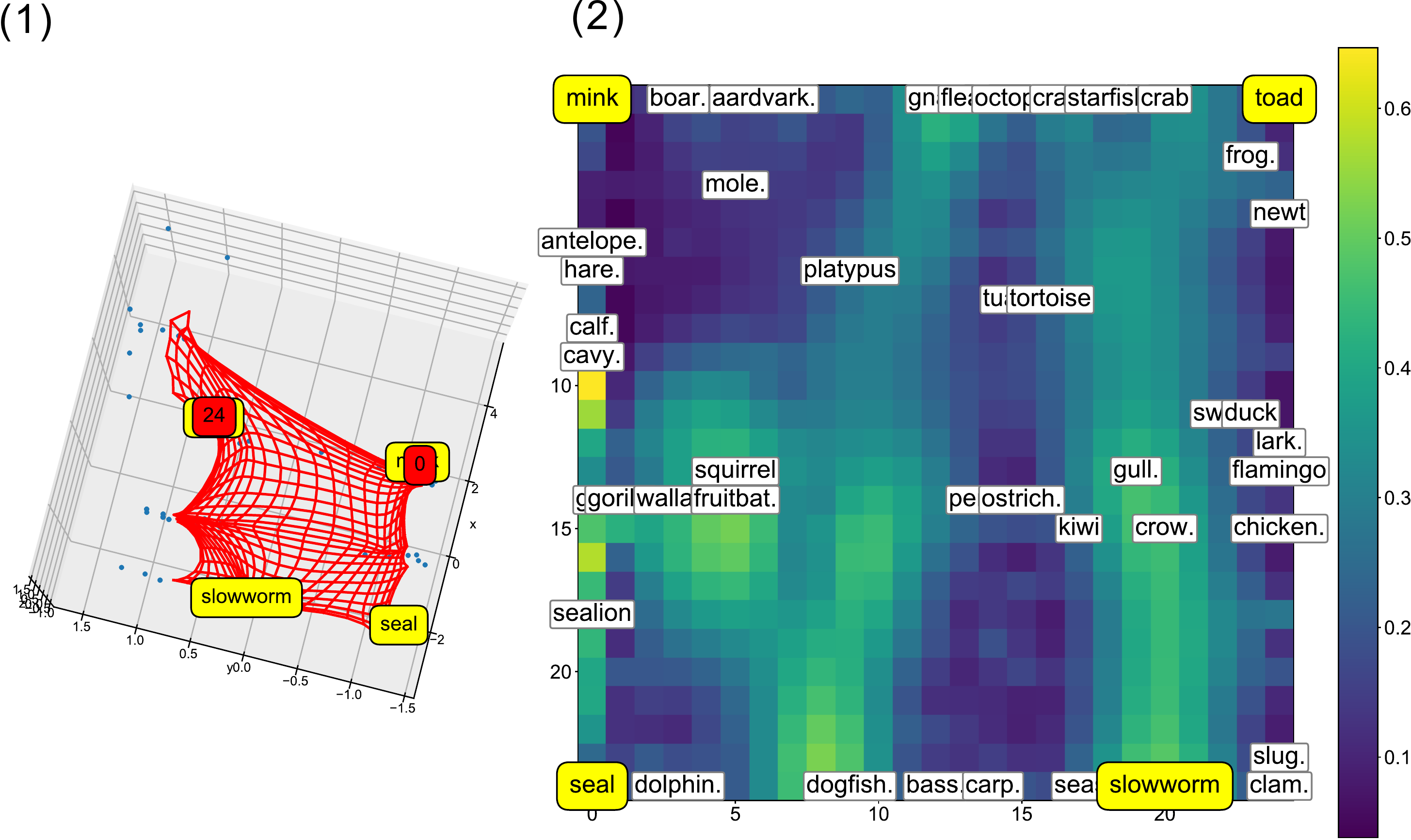}
\caption{Codebook vectors and the U-matrix of LAMA4 trained on the Zoo dataset. 
(1) {\it Codebook vectors and data in the output space}. 
Nodes 0 and 24 are shown by red labels
to illustrate the rotation of the codebook vectors.
The yellow labels indicate the location of the landmark data. 
(2) {\it Data labels on the U-matrix}. 
The white labels indicate the data projected onto the location of the U-matrix. 
The yellow labels indicate the landmark data.
The dot in the label implies that multiple labels are projected onto the same node. 
The color of the U-matrix represents the distance from the neighboring nodes. 
As the color turns yellow, the distance increases. 
}
\label{fig:zooLama4}
\end{figure}

LAMA4 was an extension of LAMA3.  
Fig~\ref{fig:zooLama4}(1) shows the codebook vectors of LAMA4. 
As all the landmark nodes were located on the edge of a square in the output space, 
the mesh grid were spread over the data with its edges close to the landmark nodes.
Fig~\ref{fig:zooLama4}(2) shows the U-matrix of LAMA4. 
The landmark data points were projected close to the edges of the U-matrix. 
The figure showed the similarities of the data, focusing on four different landmark data points.

\subsection{Evaluation of numerical error indices}
\begin{figure}[!h]
\includegraphics[width=\linewidth]{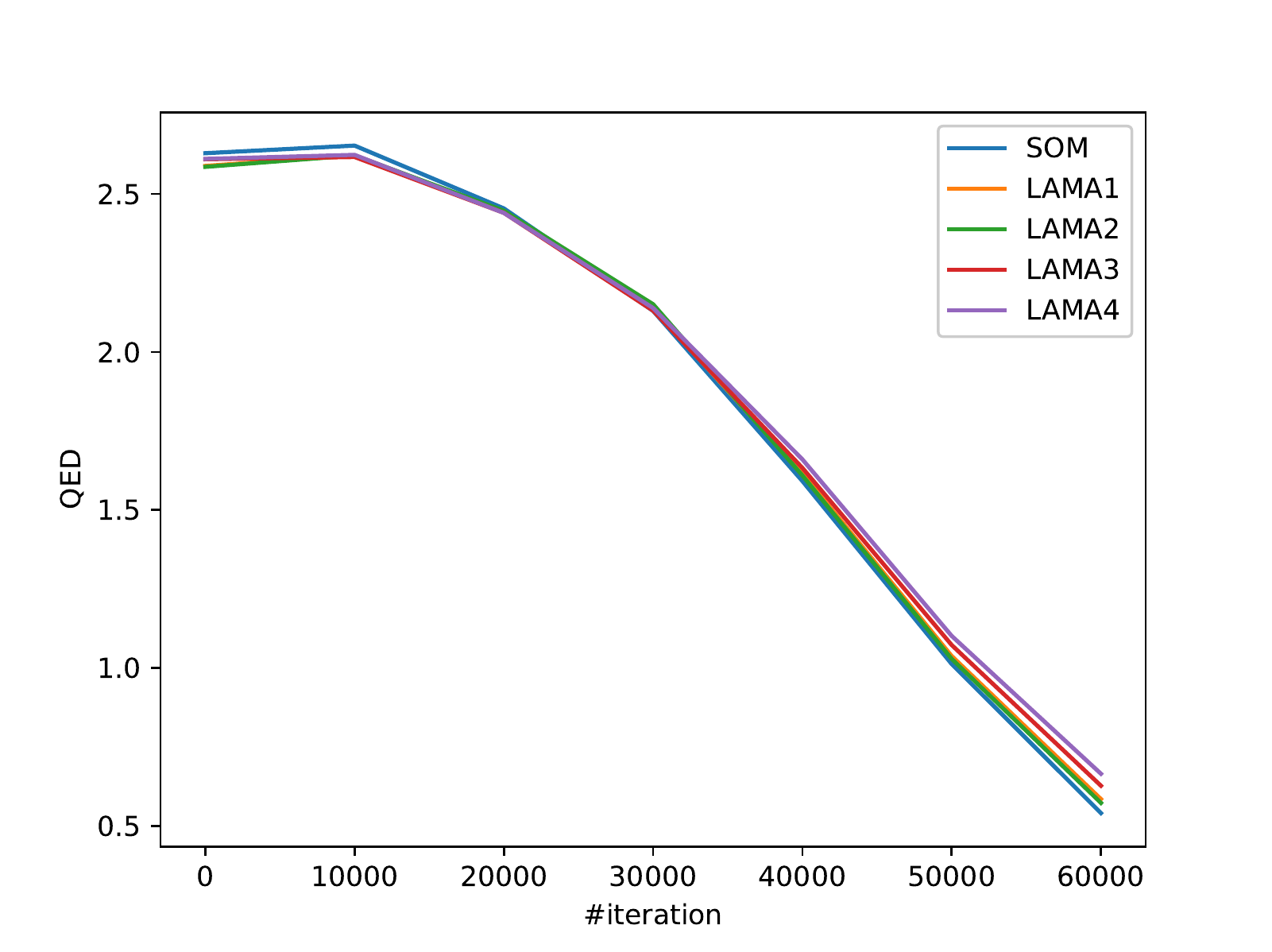}
\caption{Mean quantization error of data learned from the Zoo dataset.
}
\label{fig:ZOO_QED}
\end{figure}

\noindent
Figure~\ref{fig:ZOO_QED} shows the QED when analyzing the Zoo dataset. 
The QED of all algorithms showed similar descending curves. 
The lowest QED was achieved by SOM, followed by LAMA2, LAMA1, LAMA3, and LAMA4. 
These results imply that the codebook vectors of LAMA 
expand over the data in a similar way as SOM.

\begin{figure}[!h]
\includegraphics[width=\linewidth]{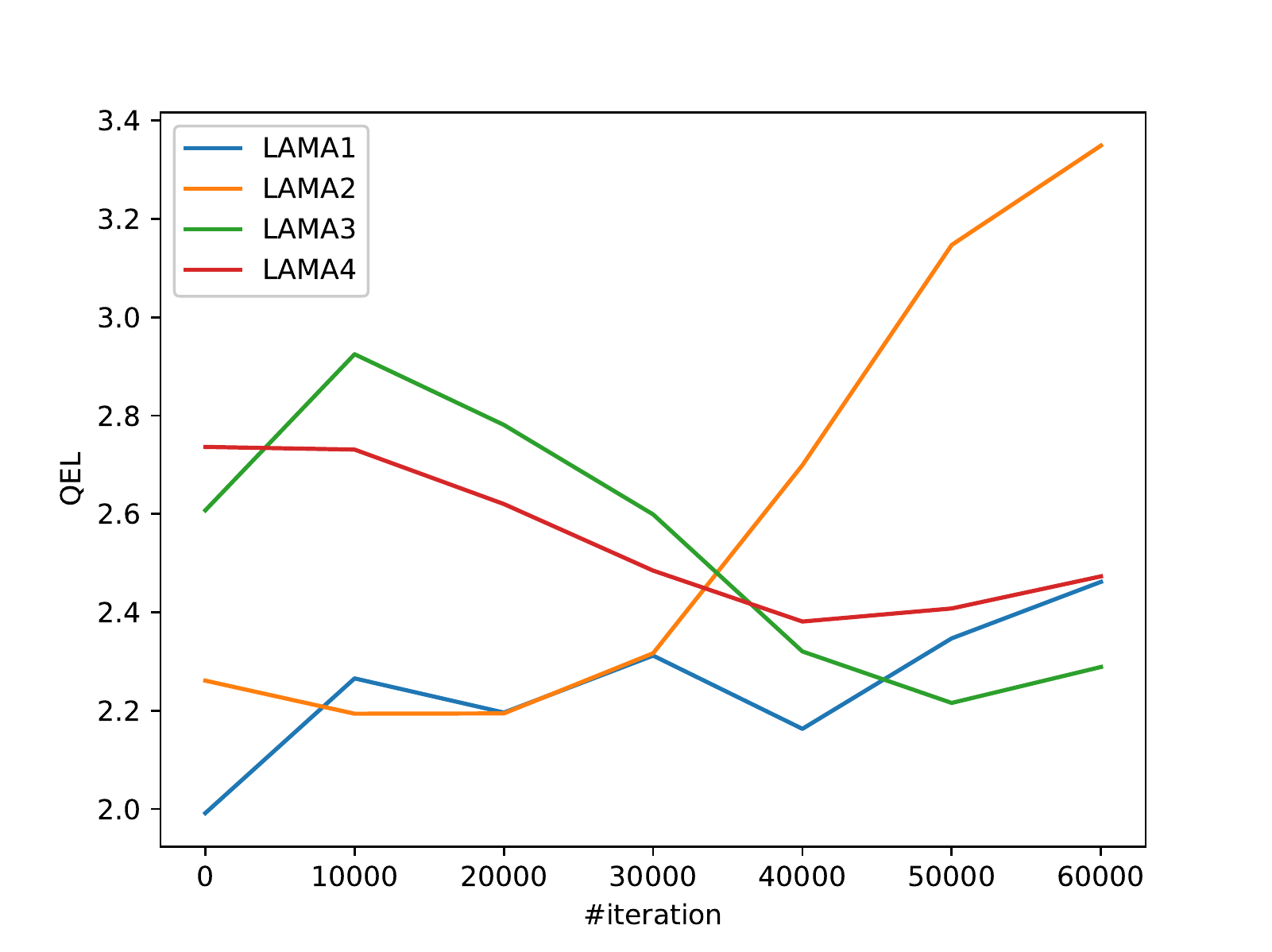}
\caption{Mean quantization error of landmark learned from the Zoo dataset.
}
\label{fig:ZOO_QEL}
\end{figure}

Figure~\ref{fig:ZOO_QEL} shows the QEL when analyzing the Zoo dataset. 
The QEL of the SOM was not analyzed because SOM did not have landmark nodes. 
Large variability can be seen between algorithms. 
The QEL of LAMA1, LAMA3, and LAMA4 remained below 2.6 at the end of the learning. 
The QEL of LAMA2 increased toward the final iteration. 
The increase of the QEL implies that the codebook vectors of landmark nodes 
are moving far from the corresponding landmark data. 
The differences of QEL in the Zoo dataset analysis may be due to the allocation of landmarks.

\begin{figure}[!h]
\includegraphics[width=\linewidth]{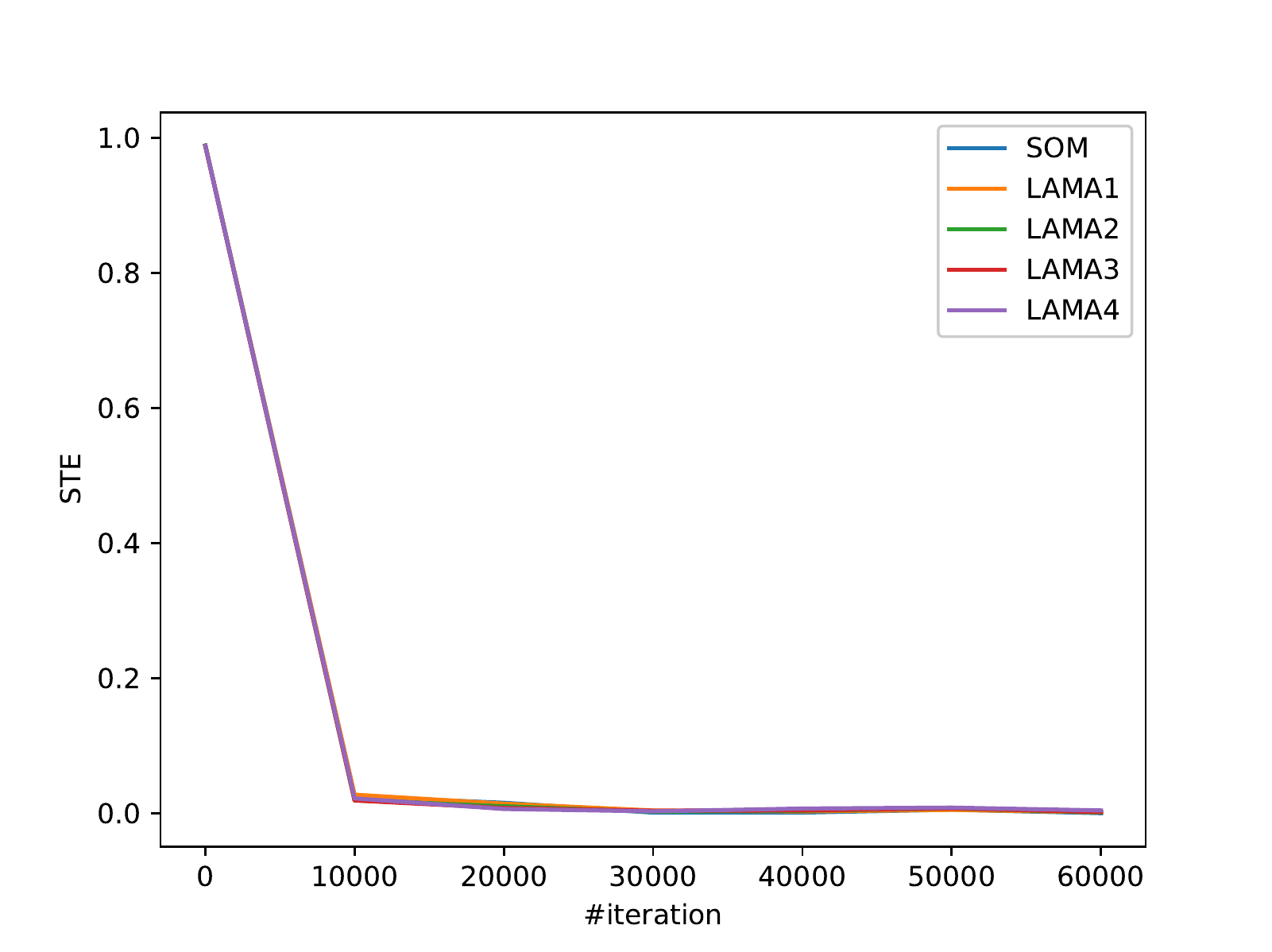}
\caption{Mean square topographic error of data learned from the Zoo dataset.
}
\label{fig:ZOO_STE}
\end{figure}

STEs are shown in Fig.~\ref{fig:ZOO_STE}. 
All algorithms show a similar tendency in terms of STE. 
They all show sudden decreases  
when the number of iterations is less than 10000. 
All algorithms maintained a low STE at the end of the learning iteration.

\clearpage

\section{Formant dataset analysis}

\subsection{Dataset and parameters}
\begin{figure}[!h]
\includegraphics[width=\linewidth]{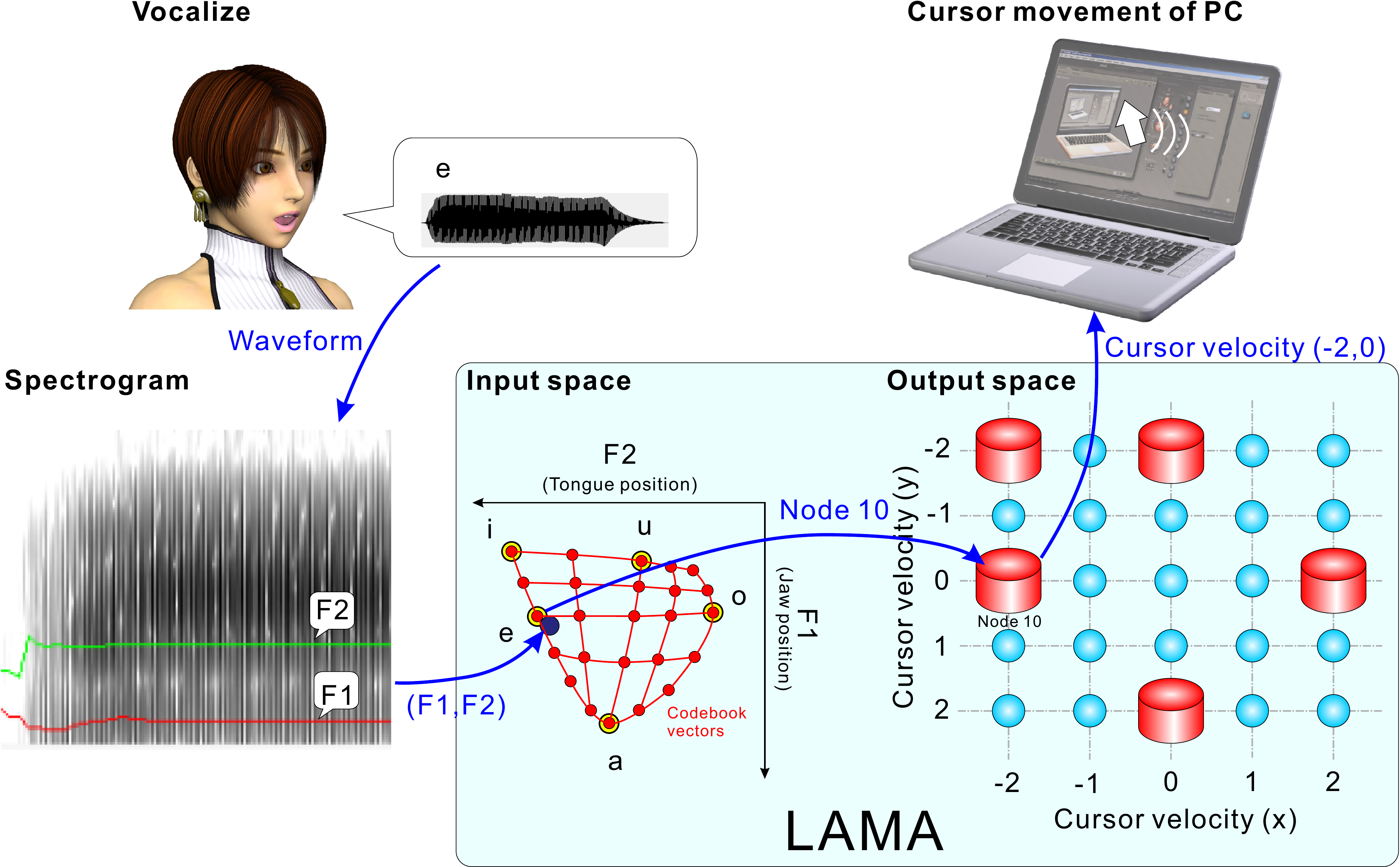}
\caption{Example of using LAMA for human-computer interaction. 
This system converts vowel sounds into vertical or horizontal movements 
considering articular movement. 
First, the sound of the Japanese vowel /e/ is recorded via a microphone. 
Then, the sound is analyzed by the spectrogram. 
From the graph, the first and second peaks in frequency are extracted, 
which are called the first and the second formants (F1, F2). 
After correcting F1 and F2 of the Japanese vowel sounds, 
they are displayed in the F1-F2 plot, 
which indicates that the jaw position is related to F2, 
while the tongue position is related to F1.
To exploit this relationship for cursor movement, 
landmarks are installed such that the tongue movement (/e/ to /o/) is 
associated with the horizontal cursor movement, 
while the jaw movement (/u/ to /a/) is related to the vertical cursor movement.
If the learning is successful, the sound data /e/ is projected onto node 10, 
and the cursor movement in a computer 
is controlled as (the velocity of x, the velocity of y)= (-2, 0). 
Thus, the LAMA works as an interface between formants and cursor movement. 
}
\label{fig:formant_dev}
\end{figure}

\noindent
This experiment provides an example of applying LAMA 
to an artificially generated formant dataset. 
Formants are important components that characterize vowels in human speech. 
They can be regarded as the peaks of the spectrogram of vowel sounds (see Fig.~\ref{fig:formant_dev}). 
The lowest peak in frequency is the first formant (F1) 
and the next one is the second formant (F2). 
In practice, the peaks are extracted from the envelope of the spectrogram by linear predictive coding.
For example, the mean first and second formants (F1, F2) of 
the five Japanese vowels (/a/, /i/, /u/, /e/, and /o/) 
are (850, 1610), (240, 2400), (300, 1390), (390, 2300), and (360, 640), respectively \cite{nakagawa1980differences}. 
Here, /$\cdot$/ indicates phonetic transcription (broad transcription) of the speech sounds. 
As the changes in F1 and F2 are related to the movement of the tongue and the jaw, respectively, 
F1 and F2 can be used for pointing a cursor on a computer. 
A pointing device using formants has been proposed \cite{uemi2013study}. 
Such a device can be extended by 
a projection between the formants and the cursor movement using LAMA. 
This demonstration aims to explain how to apply the LAMA to HCI.

To reveal the difference between the learning properties of LAMA and SOM, 
they were trained on an artificially generated formant dataset, 
which is provided in Supplementary Material 1. 
The parameters, including the landmarks, are presented in Table~\ref{table:paramFormant}. 
The mean F1 and F2 for each Japanese vowel were designated as the landmark data points.
The landmark nodes were arranged in the output space 
such that the horizontal cursor movement was related to the tongue movement (/e/,/o/) 
while the vertical cursor movement was associated with the jaw movement (/u/,/a/).  
Note that the landmark data was not selected from the given dataset.

\begin{table}[!ht]
\centering
\caption{
{Parameters of SOM and LAMA used for the artificial formant dataset analysis.}}
\begin{tabular}{c|ccccc}
  	\hline
    Algorithm 			& SOM 			& LAMA \\
    \hline
    $\mathrm{K}_x$ 				& 10 			& 10 			\\
    $\mathrm{K}_y$ 				& 10 			& 10 			\\
    $t_\mathrm{max}$ 			& 60000 		& 60000 		\\
    $a_\mathrm{max}$ 			& 0.3 			& 0.3 			\\
    $a_\mathrm{min}$ 			& 0.1 			& 0.05 			\\
    $\tau_a$			& $t_\mathrm{max}/3-1$	& $t_\mathrm{max}/3-1$	\\
    $\sigma_\mathrm{max}$		& 4 			& 4				\\
    $\sigma_\mathrm{min}$		& 0.3			& 0.4			\\
    $\tau_\sigma$		& $t_\mathrm{max}/3-1$	& $t_\mathrm{max}/3-1$	\\
    $b_\mathrm{max}$ 			& --			& 0.3 			\\
    $b_\mathrm{min}$ 			& -- 			& 0.08 			\\
    $\tau_b$			& --			& $t_\mathrm{max}/3-1$	\\
    $t_\mathrm{center}$ 		& --	 		& 30000 		\\
    $\rho_{b}$			& --			& 15000			\\
    $\rho_\mathrm{max}$		& --			& 2			\\
    $\rho_\mathrm{min}$		& --			& 0.8		\\
    $\tau_\rho$			& --			& $t_\mathrm{max}/3-1$	\\
    $p_\mathrm{th}$			& --			& 0.1			\\
    Landmark $(vowel:k)$	& --			& a:94, i:0, u:4, e:41, o:49 \\
    \hline
  \end{tabular}
\begin{flushleft} 
\end{flushleft}
\label{table:paramFormant}
\end{table}

Moreover, QED, QEL, and STE were calculated 
for the artificial formant dataset 
as well as the Zoo dataset analysis.

\subsection{Visualization of learning properties}
\begin{figure}[!h]
\includegraphics[width=\linewidth]{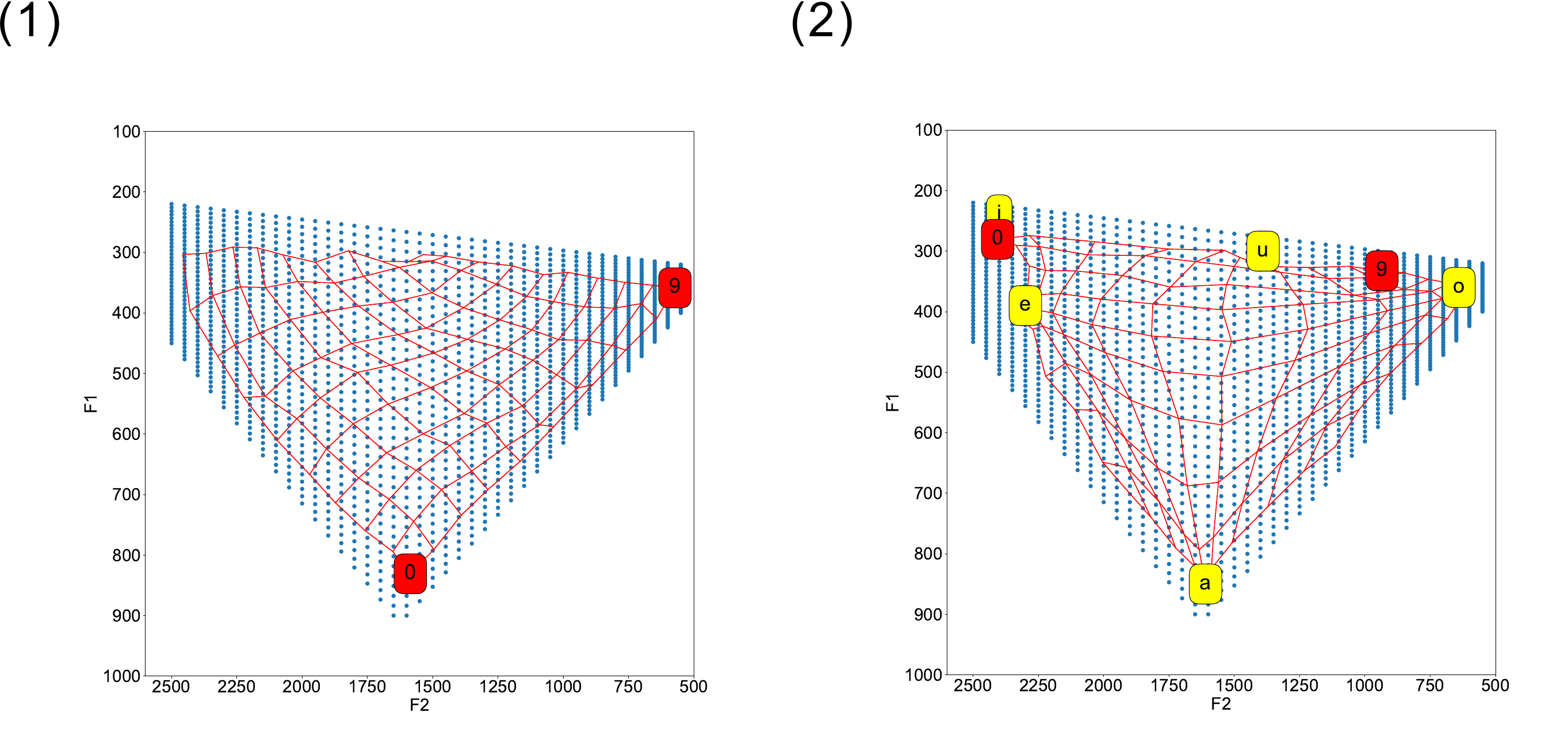}
\caption{Codebook vectors of SOM and LAMA trained on the artificial formant dataset. 
(1) {\it Codebook vectors of the SOM and data in the input space}. 
The blue dots represent the data points. 
The crossing points of the red lines indicate the codebook vectors of the SOM.  
(2) {\it Codebook vectors of LAMA and data in the input space}.  
The yellow labels indicate the landmark data.
}
\label{fig:formantSomLama}
\end{figure}

LAMA provided an intuitive relationship between the articulator movement 
and the computer cursor movement. 
Fig.~\ref{fig:formantSomLama}(1) shows the codebook vectors of the SOM.
The red mesh grid composed of codebook vectors was smoothly spread over the given data points. 
The horizontal movement of the tongue (movement between /e/ and /o/) 
was not related to horizontal movement in the output space. 
In addition, jaw movement (/u/, /a/) was not fit to vertical movement in the output space. 
The mesh grid might be rotated because the codebook vectors were randomly initialized and 
no constraints were given for learning. 
Fig.~\ref{fig:formantSomLama}(2) shows the codebook vectors of LAMA.
The red mesh grid composed of codebook vectors was spread over the given data points. 
The mesh grid had a line between /e/ and /o/, 
which corresponded to the horizontal movement of the tongue. 
This means that the horizontal movement of the tongue 
is directly reflected by the horizontal movement in the output space. 
Furthermore, the mesh grid contained a line between /u/ and /a/, 
which was related to the vertical movement of the jaw. 
This implies that the vertical movement of the jaw is retained in the output space.
Thus, LAMA in this example provides a projection that maintains the articulator movement in the output space.

\subsection{Evaluation of numerical error indices}

\begin{figure}[!h]
\includegraphics[width=\linewidth]{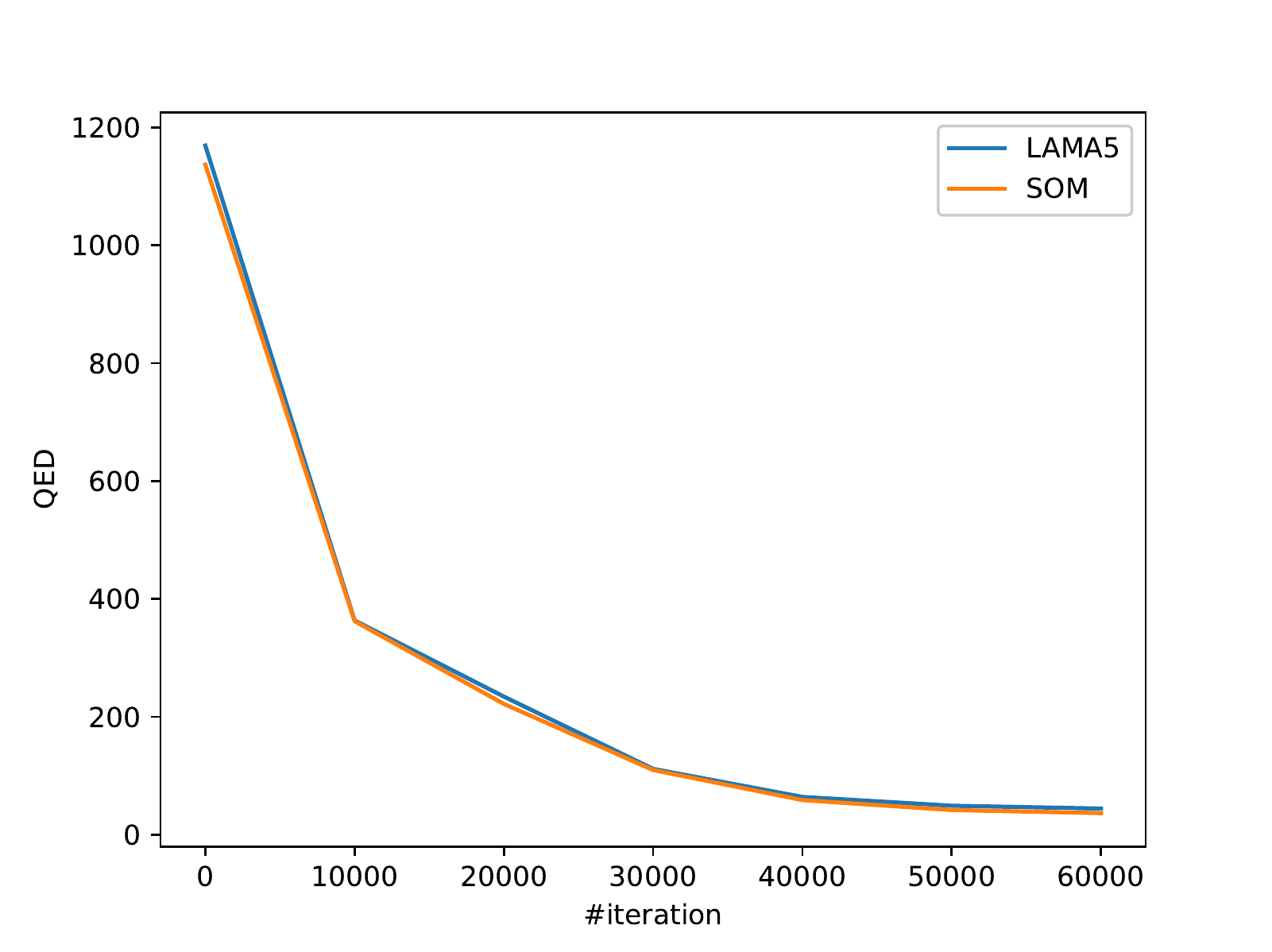}
\caption{Mean quantization error of data learned from the artificial formant dataset. 
}
\label{fig:Formant_QED}
\end{figure}

\begin{figure}[!h]
\includegraphics[width=\linewidth]{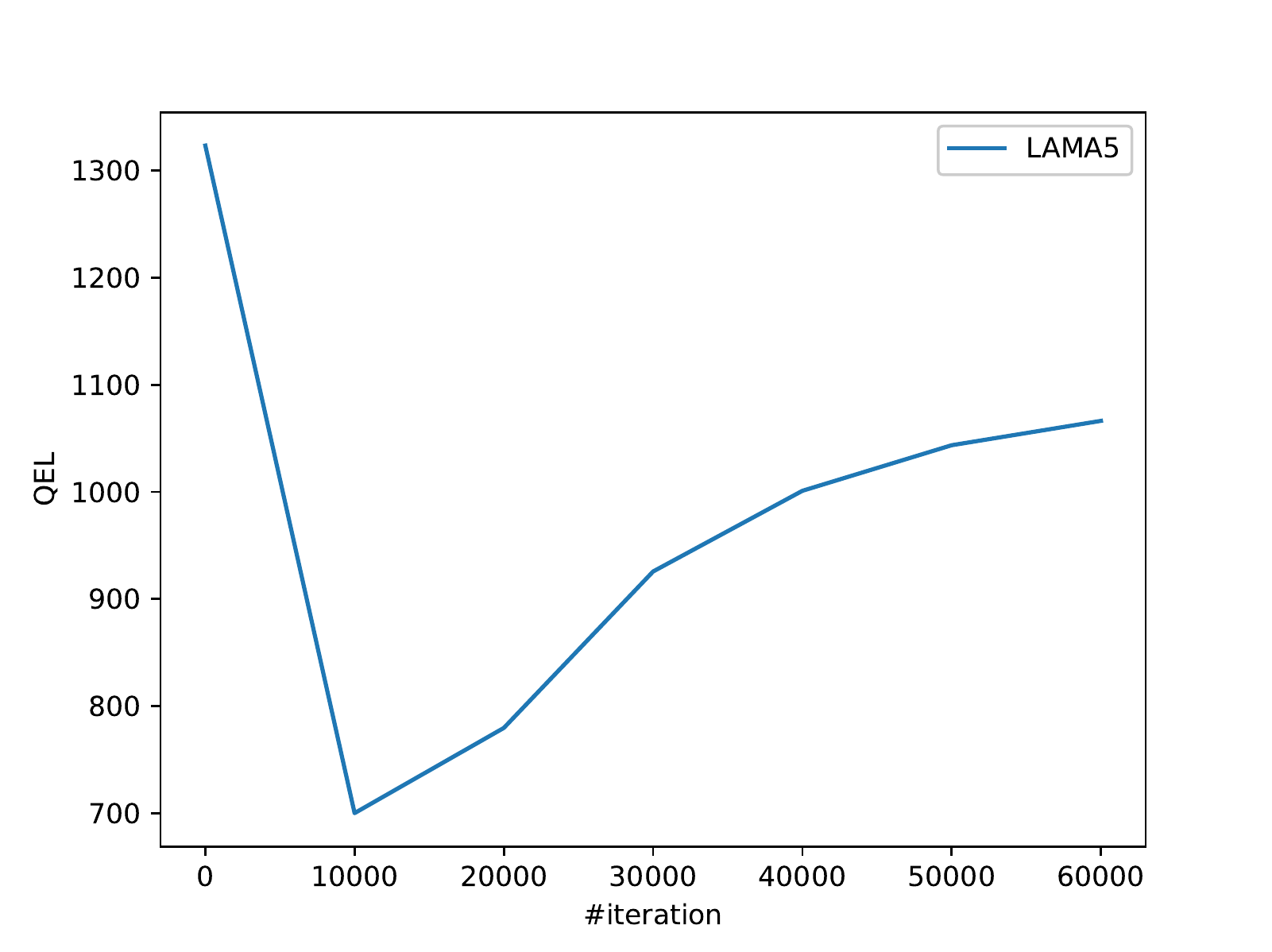}
\caption{Mean quantization error of landmark learned from the artificial formant dataset. 
}
\label{fig:Formant_QEL}
\end{figure}

\begin{figure}[!h]
\includegraphics[width=\linewidth]{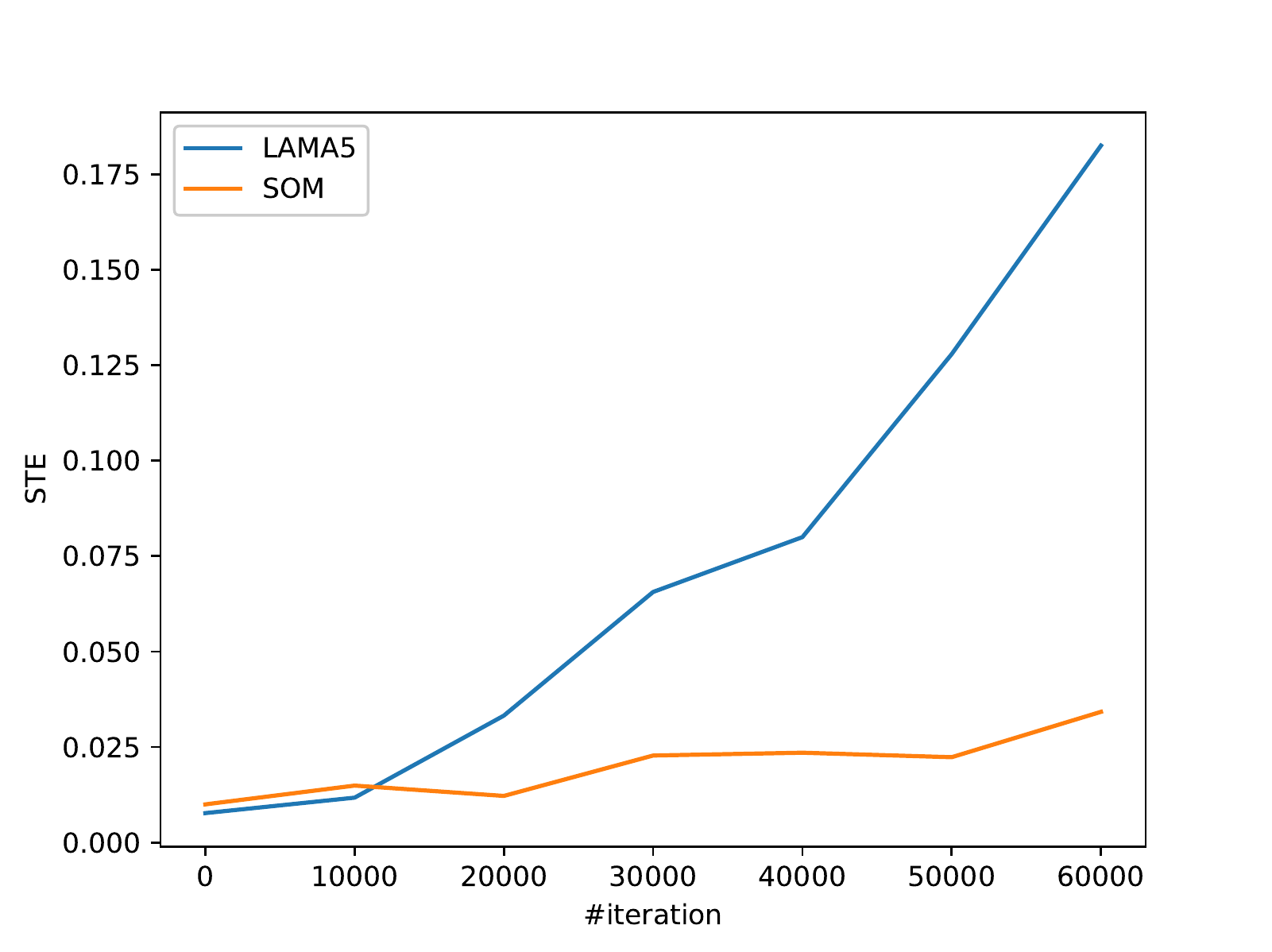}
\caption{Mean square topographic error of data learned from the artificial formant dataset. 
}
\label{fig:Formant_STE}
\end{figure}

Figure~\ref{fig:Formant_QED} shows the QED 
when analyzing the artificial formant dataset. 
The QEDs of both SOM and LAMA decrease monotonically. 
This implies that LAMA5 preserves the learning properties of SOM. 
Figure~\ref{fig:Formant_QEL} represents the QEL of LAMA5. 
It increased toward the end of learning. 
This result implies that landmark nodes were gradually far from the 
landmark data during learning. 
Figure~\ref{fig:Formant_STE} shows the STE.
Both SOM and LAMA5 show an increase of the STE. 
The STE of SOM maintained a low level (below 0.05) compared to that of LAMA5.  
This result implies that topological errors occasionally occur
due to the settings of landmarks or parameters.

\clearpage

\section{Discussion}
\noindent
Herein, LAMA was proposed, 
which provides the user-intended nonlinear projection 
by designating several landmark nodes. 
The experiment demonstrated that LAMA had its own learning 
properties compared with the SOM. 
In the analysis of the Zoo dataset, 
LAMA demonstrated that it provided varieties of landmark-oriented visualization. 
The QED of the SOM and LAMA showed similar tendencies and became small 
in the final learning condition, 
implying that the codebook vectors of LAMA 
spread over data in the same way as a SOM, 
while imposing constraints of some nodes and data.
This study also demonstrated that the movement of the articulator 
was successfully maintained in the output space using LAMA.  
These results suggest that 
LAMA can be used for designing new HCI devices, 
in addition to data mining.

LAMA can describe the relationship between the input and 
output spaces by designating several landmark nodes. 
Thus, it is feasible to apply LAMA 
to HCI by converting 
biosignals into computer commands. 
For example, limited finger movements of people 
who have paralysis can be converted into, for example, cursor movements on a computer. 
Other biosignals, such as EEGs, 
electromyographs, and electrooculograms, 
can be translated into two-dimensional output space 
by applying a relationship between the input and output spaces.

In addition to exploiting biosignals, 
LAMA can be used to summarize or explore data. 
For example, LAMA can contribute to the design of 
a new recommendation system of products using 
popular products as landmarks. 
Such a system can help users search for products 
by some designated popular products. 
Once the product data are represented 
by features, as in the case of the Zoo dataset, 
LAMA can show the relationship between 
the given data with landmark nodes and the other data. 
However, the current U-matrix may not be suitable 
for use in a recommendation system 
because some data names are projected into the same node. 
Moreover, each item in the U-matrix is 
too small to visualize the image of products. 
A new visualization method is required for  
recommendation system to be more practical.

LAMA learns codebook vectors, 
attempting to make some codebook vectors fit the assigned landmark data 
and retaining the learning properties of the SOM. 
However, these two objectives cannot always be completely achieved simultaneously. 
If the landmark nodes learn stronger parameters in the landmark-driven phase, 
the codebook vectors of the landmark nodes may fit the landmark data 
with the mesh grid distorted. 
Such distortion will be large as the designation of the landmark nodes 
is different from that of the codebook vectors learned by the SOM. 
On the other hand, if stronger parameters are used in the data-driven phase, 
the learning properties are close to those of the SOM and 
the landmark nodes do not receive much consideration. 
To achieve both objectives, 
the parameters should be adjusted 
by confirming the visualization of the codebook vectors.

The learning parameters of LAMA are not determined automatically, 
but are manually adjusted by estimation 
from the visualization of the codebook vectors. 
Missetting of the parameters may cause the mesh grid of 
codebook vectors to be twisted or wrinkled.
However, there is no strict rule for setting the parameters. 
In this study, parameters shared with the SOM were adjusted first, 
then all parameters were modified. 
By designing proper learning rates in the data-driven phase, 
the codebook vectors were spread over the given data. 
Second, the landmark nodes were arranged, and the learning rates of the landmark-driven phase and 
the update rate of the alternating update method were then manipulated 
such that the codebook vectors fit the corresponding landmark data points 
with the smooth mesh grid.

Setting up of landmark nodes should be based on the assumption or objective for a projection. 
For example, the landmark nodes for the F1-F2 projection in this study were prepared 
such that the articular movement is compatible with the movement in the output space. 
It would be preferable to assign landmark data to the projected node of the SOM 
because it does not require LAMA for difficult fitting. 
As the user-intended projection of LAMA is far from the projection of the SOM, 
LAMA needs to modify the mesh grid considerably, which causes wrinkles on the grid. 
Thus, it would be preferable to install the landmark nodes 
by visualizing the data and codebook vectors of the SOM.

The arrangement of codebook vectors in the input space may rotate, 
depending on the way the landmarks are set. 
If no landmark is installed (SOM), 
its codebook vectors can be rotated, depending on the initial values.
If a landmark is set at the center of the nodes (LAMA1), 
the codebook vectors are allowed to be rotated, centering the landmark. 
When two landmarks are designated (LAMA2), 
the codebook vectors can rotate, centering the axis between two landmarks. 
In the same way, the degree of freedom for LAMA depends on the
input data and the number and location of landmarks installed.

This study employed QED, QEL, and STE to evaluate the learning properties of SOM and LAMA numerically. 
The optimization problem of LAMA can be explained in that 
it minimizes the QED and QEL simultaneously. 
It successfully reduced the QED while keeping the QEL low in this study. 
Toward better evaluation of LAMA, 
more comprehensive error indices, 
such as combined indices of QED, QEL, and STE, 
would be helpful. 
The STE was successfully reduced in the Zoo dataset analysis. 
However, the STE increased when LAMA5 was applied to the artificial formant analysis. 
This implies that the learning problem is difficult due to landmark settings, 
and requires several learning trials to obtain the user-intended nonlinear projection. 
Additional learning phases, such as a smoothing phase, may contribute to further reduction of the STE.

In future studies, LAMA can be improved by applying  
advanced algorithms proposed for the SOM or the generative topographic mapping (GTM). 
For example, the online learning method 
can be extended to the batch learning method \cite{kohonen2001}. 
Rauber et al.~proposed a growing hierarchical SOM  
that has a dynamic architecture for improving its representation capacity \cite{Rauber2002}. 
Such a growing hierarchical model for LAMA may 
reduce difficulties in setting the parameters. 
In addition, there remains scope for the development of tensor analysis, 
such as that used in the tensor SOM and the tensor GTM \cite{Iwasaki2016}. 
Furthermore, LAMA can be cascaded similarly to the SOM of SOMs \cite{Furukawa2009}. 
LAMA may be modeled by a continuous model such as the GTM \cite{Bishop1998}. 
The SOM and GTM have been evaluated on the presence of missing data \cite{Vatanen2015}. 
The learning properties of LAMA against such data should be clarified in future studies.

\section{Conclusion}
In this paper, LAMA was introduced, 
which provides the user-intended nonlinear projection 
from input data in an input space to a discrete output space 
by setting several landmark nodes. 
This study demonstrated that 
LAMA could provide unique data visualization, 
for example, landmark-oriented visualization. 
In addition, an example was presented to demonstrate projection 
using an artificial formant dataset, 
which enable us to design an HCI application. 
LAMA can be applied to develop new HCI devices or 
recommendation systems.

\section*{Acknowledgements}

This research did not receive grants from 
funding agencies in the public, commercial, or not-for-profit sectors.


\bibliography{mybibfile}

\begin{thebibliography}{10}
\expandafter\ifx\csname url\endcsname\relax
  \def\url#1{\texttt{#1}}\fi
\expandafter\ifx\csname urlprefix\endcsname\relax\def\urlprefix{URL }\fi
\expandafter\ifx\csname href\endcsname\relax
  \def\href#1#2{#2} \def\path#1{#1}\fi

\bibitem{kohonen2001}
T.~Kohonen, Self-organizing maps, 3rd Edition, Springer series in information
  sciences, 30, Springer, 2001.

\bibitem{Kohonen2013}
T.~Kohonen, {Essentials of the self-organizing map}, Neural Networks 37 (2013)
  52--65.

\bibitem{kraaijveld1995nonlinear}
M.~A. Kraaijveld, J.~Mao, A.~K. Jain, A nonlinear projection method based on
  kohonen's topology preserving maps, IEEE Transactions on neural networks
  6~(3) (1995) 548--559.

\bibitem{Uriarte2005}
E.~A. Uriarte, F.~D. Mart{\'{i}}n, {Topology preservation in SOM},
  International Journal of Applied Mathematics and Computer Sciences 1~(1)
  (2005) 19--22.

\bibitem{ultsch1990kohonen}
A.~Ultsch, Kohonen's self organizing feature maps for exploratory data
  analysis, Proc. INNC90 (1990) 305--308.

\bibitem{Ultsch2003}
A.~Ultsch, {U * -matrix : A tool to visualize clusters in high dimensional
  data}, Tech. Rep.~36 (2003).

\bibitem{Vesanto2000}
J.~Vesanto, E.~Alhoniemi, {Clustering of the self-organizing map}, IEEE
  Transactions on Neural Networks 11~(3) (2000) 586--600.

\bibitem{Caldas2018}
R.~Caldas, D.~R{\'{a}}tiva, F.~{Buarque de Lima Neto}, {Clustering of
  self-organizing maps as a means to support gait kinematics analysis and
  symmetry evaluation}, Medical Engineering and Physics~(62) (2018) 46--52.

\bibitem{Sulkava2015}
M.~Sulkava, A.~M. Sepponen, M.~Yli-Heikkil{\"{a}}, A.~Latukka, {Clustering of
  the self-organizing map reveals profiles of farm profitability and upscaling
  weights}, Neurocomputing 147~(1) (2015) 197--206.

\bibitem{Belkhiri2018}
L.~Belkhiri, L.~Mouni, A.~Tiri, T.~S. Narany, R.~Nouibet, {Spatial analysis of
  groundwater quality using self-organizing maps}, Groundwater for Sustainable
  Development 7~(June 2017) (2018) 121--132.

\bibitem{Ultsch2005}
A.~Ultsch, F.~M{\"{o}}rchen, {ESOM-maps: Tools for clustering, visualization,
  and classification with emergent SOM}, Tech. rep. (2005).

\bibitem{Al-Ketbi2013}
O.~Al-Ketbi, M.~Conrad, {Supervised ANN vs. unsupervised SOM to classify EEG
  data for BCI: Why can GMDH do better?}, International Journal of Computer
  Applications 74~(4) (2013) 37--44.

\bibitem{Lawrence1997}
S.~Lawrence, C.~L. Giles, A.~C. Tsoi, A.~D. Back, {Face recognition: A
  convolutional neural-network approach}, IEEE Transactions on Neural Networks
  8~(1) (1997) 98--113.

\bibitem{Coleca2015}
F.~Coleca, A.~State, S.~Klement, E.~Barth, T.~Martinetz, {Self-organizing maps
  for hand and full body tracking}, Neurocomputing 147~(1) (2015) 174--184.

\bibitem{Chen2016}
H.~Chen, Q.~Gao, T.~Feng, Y.~Liu, X.~Xiao, {Body falling gesture recognition
  based on SOM and triaxial acceleration information}, in: Proceedings of the
  2016 international conference on commpunications, information management and
  network security, Vol.~47, 2016, pp. 70--73.

\bibitem{Yu2015}
H.~Yu, F.~Khan, V.~Garaniya, {Risk-based fault detection using self-organizing
  map}, Reliability Engineering and System Safety 139 (2015) 82--96.

\bibitem{Liu2005}
H.~Liu, J.~Wang, C.~Zheng, {Using self-organizing map for mental tasks
  classification in brain-computer interface}, Advances in Neural Networks ISNN
  Second International Symposium on Neural Networks 2 (2005) 327--332.

\bibitem{Khosrowabadi2010}
R.~Khosrowabadi, H.~C. Quek, A.~Wahab, K.~K. Ang, {EEG-based emotion
  recognition using self-organizing map for boundary detection}, 2010 20th
  International Conference on Pattern Recognition (2010) 4242--4245.

\bibitem{Majumder2014}
A.~Majumder, L.~Behera, V.~K. Subramanian, {Emotion recognition from geometric
  facial features using self-organizing map}, Pattern Recognition 47~(3) (2014)
  1282--1293.

\bibitem{Selmanaj2017}
D.~Selmanaj, M.~Corno, S.~M. Savaresi, {Hazard detection for motorcycles via
  accelerometers: A self-organizing map approach}, IEEE Transactions on
  Cybernetics 47~(11) (2017) 3609--3620.

\bibitem{kohonen1988neural}
T.~Kohonen, The 'neural' phonetic typewriter, Computer 21~(3) (1988) 11--22.

\bibitem{fessant2001comparison}
F.~Fessant, P.~Aknin, L.~Oukhellou, S.~Midenet, Comparison of supervised
  self-organizing maps using euclidian or mahalanobis distance in
  classification context, in: International Work-Conference on Artificial
  Neural Networks, Springer, 2001, pp. 637--644.

\bibitem{goren2000supervised}
D.~Goren-Bar, T.~Kuflik, D.~Lev, Supervised learning for automatic
  classification of documents using self-organizing maps, in: DELOS Workshop:
  Information Seeking, Searching and Querying in Digital Libraries, 2000.

\bibitem{Kohonen1998}
T.~Kohonen, P.~Somervuo, {Self-organizing maps of symbol strings},
  Neurocomputing 21~(1-3) (1998) 19--30.

\bibitem{Hagenbuchner2005}
M.~Hagenbuchner, A.~C. Tsoi, {A supervised training algorithm for
  self-organizing maps for structures}, Pattern Recognition Letters 26~(12)
  (2005) 1874--1884.

\bibitem{Shen2010}
F.~Shen, H.~Yu, Y.~Kamiya, O.~Hasegawa, {An online incremental semi-supervised
  learning method}, Journal of Advanced Computational Intelligence and
  Intelligent Informatics 14~(6) (2010) 593--605.

\bibitem{Herrmann2007}
L.~Herrmann, A.~Ultsch, Label propagation for semi-supervised learning in
  self-organizing maps, in: In The 6th International Workshop on
  Self-Organizing Maps (WSOM 2007), 2007, pp. 3--6.

\bibitem{Meschino2015}
G.~J. Meschino, D.~S. Comas, V.~L. Ballarin, A.~G. Scandurra, L.~I. Passoni,
  {Automatic design of interpretable fuzzy predicate systems for clustering
  using self-organizing maps}, Neurocomputing 147~(1) (2015) 47--59.

\bibitem{nakagawa1980differences}
S.~Nakagawa, H.~Shirakata, M.~Yamao, T.~Sakai, Differences in feature
  parameters of japanese vowels with sex and age, Sutudia Phonologica XIV
  (1980) 33--52.

\bibitem{uemi2013study}
N.~Uemi, A study of a human interface device controlled by formant frequencies
  for the disabled, in: International Conference on Cross-Cultural Design,
  Springer, 2013, pp. 340--345.

\bibitem{Rauber2002}
A.~Rauber, D.~Merkl, M.~Dittenbach, {The growing hierarchical self-organizing
  map: Exploratory analysis of high-dimensional data}, IEEE Transactions on
  Neural Networks 13~(6) (2002) 1331--1341.

\bibitem{Iwasaki2016}
T.~Iwasaki, T.~Furukawa, {Tensor SOM and tensor GTM: Nonlinear tensor analysis
  by topographic mappings}, Neural Networks 77 (2016) 107--125.

\bibitem{Furukawa2009}
T.~Furukawa, {SOM of SOMs}, Neural Networks 22~(4) (2009) 463--478.

\bibitem{Bishop1998}
C.~M. Bishop, M.~Svens{\'{e}}n, C.~K.~I. Williams, {GTM: The Generative
  Topographic Mapping}, Neural Computation 10~(1) (1998) 215--234.

\bibitem{Vatanen2015}
T.~Vatanen, M.~Osmala, T.~Raiko, K.~Lagus, M.~Sysi-Aho, M.~Ore{\v{s}}i{\v{c}},
  T.~Honkela, H.~L{\"{a}}hdesm{\"{a}}ki, {Self-organization and missing values
  in SOM and GTM}, Neurocomputing 147~(1) (2015) 60--70.

\end{thebibliography}

\end{document}